%% file: main.tex
\documentclass{article}

\usepackage{microtype}
\usepackage{graphicx}
\usepackage{subfigure}
\usepackage{booktabs} 

\usepackage{multirow, makecell}

\usepackage{hyperref}

\usepackage[accepted]{icml2024}

\usepackage{amsmath}
\usepackage{amssymb}
\usepackage{mathtools}
\usepackage{amsthm}

\usepackage[capitalize,noabbrev]{cleveref}

\usepackage{enumitem}

\usepackage{xspace}

\usepackage{xcolor}
\newcommand{\rt}[1]{\textcolor{red}{#1}\xspace}
\definecolor{darkergray}{RGB}{30,30,30}
\newcommand{\estimated}[1]{\textcolor{darkergray}{\textit{#1}}}

\definecolor{darkred}{RGB}{140,0,0}

\theoremstyle{plain}

\theoremstyle{definition}

\theoremstyle{remark}

\usepackage[textsize=tiny]{todonotes}

\usepackage{comment}
\usepackage{bm}
\usepackage{leftindex}

\icmltitlerunning{Differentiable Weightless Neural Networks}

\begin{document}

\twocolumn[
\icmltitle{Differentiable Weightless Neural Networks}



\icmlsetsymbol{equal}{*}

\begin{icmlauthorlist}
\icmlauthor{Alan T. L. Bacellar}{equal,ufrj}
\icmlauthor{Zachary Susskind}{equal,ut}
\icmlauthor{Mauricio Breternitz Jr.}{iscte}
\icmlauthor{Eugene John}{utsa}
\icmlauthor{Lizy K. John}{ut}
\icmlauthor{Priscila M. V. Lima}{ufrj}
\icmlauthor{Felipe M. G. França}{it}
\end{icmlauthorlist}

\icmlaffiliation{ufrj}{Federal University of Rio de Janeiro, Brazil}
\icmlaffiliation{ut}{The University of Texas at Austin, USA}
\icmlaffiliation{utsa}{The University of Texas at San Antonio, USA}
\icmlaffiliation{iscte}{ISCTE - Instituto Universitario de Lisboa, Lisbon, Portugal}
\icmlaffiliation{it}{Instituto de Telecomunicações, Porto, Portugal}

\icmlcorrespondingauthor{Alan T. L. Bacellar}{alanbacellar@poli.ufrj.br}
\icmlcorrespondingauthor{Zachary Susskind}{zsusskind@utexas.edu}

\icmlkeywords{Weightless Neural Networks, Binary Neural Networks, Multiplication-Free Models, Edge Inference}

\vskip 0.3in
]



\printAffiliationsAndNotice{\icmlEqualContribution} 

\input{00_abstract2}
\input{01_introduction3}
\input{02_background}

\input{03_methodology}
\input{04_experiments}
\input{05_conclusion}

\section*{Acknowledgements}
This research was supported in part by Semiconductor Research Corporation (SRC) Task 3148.001,  National Science Foundation (NSF) Grant \#2326894, NVIDIA Applied Research Accelerator Program Grant, and by FCT/MCTES through national funds and, when applicable, co-funded by EU funds under the project UIDB 50008/2020, and by Next Generation EU through PRR Project Route 25 (C645463824-00000063). Any opinions, findings, conclusions, or recommendations are those of the authors and not of the funding agencies.

\section*{Impact Statement}
This paper presents work to advance the field of Machine Learning, particularly on low-power, low-latency, and low-cost devices. There are many potential societal consequences of our work, including making tiny intelligent models with high accuracies for edge applications such as healthcare monitoring.
Works that seek to lower the adoption cost and improve the energy efficiency of machine learning, such as this one, additionally have the inherent potential to improve global equality of access to technology, particularly as it impacts communities with limited financial resources and unreliable access to electricity.

\bibliographystyle{icml2024}
\bibliography{bibliography}

\input{A0_appendix}

\end{document}

%% file: 00_abstract2.tex
\begin{abstract}

We introduce the \underline{D}ifferentiable \underline{W}eightless Neural \underline{N}etwork (DWN), a model based on interconnected lookup tables.
Training of DWNs is enabled by a novel Extended Finite Difference technique for approximate differentiation of binary values.
We propose Learnable Mapping, Learnable Reduction, and Spectral Regularization to further improve the accuracy and efficiency of these models.
We evaluate DWNs in three edge computing contexts: (1) an FPGA-based hardware accelerator, where they demonstrate superior latency, throughput, energy efficiency, and model area compared to state-of-the-art solutions, (2) a low-power microcontroller, where they achieve preferable accuracy to XGBoost while subject to stringent memory constraints, and (3) ultra-low-cost chips, where they consistently outperform small models in both accuracy and projected hardware area.
DWNs also compare favorably against leading approaches for tabular datasets, with higher average rank.
Overall, our work positions DWNs as a pioneering solution for edge-compatible high-throughput neural networks.
\href{https://github.com/alanbacellar/DWN}{https://github.com/alanbacellar/DWN}

\end{abstract}

%% file: 01_introduction3.tex
\vspace{-6mm}
\section{Introduction}
Despite the rapid advancement of deep learning, optimizing computational efficiency, especially during inference, remains a critical challenge. Efforts to mitigate computational demands have led to innovations in model pruning \cite{prune1, prune2, prune3}, quantization \cite{quantization1, quantization2, quantization3}, and sparse neural networks \cite{sparse1, sparse2, sparse3}.
However, these approaches do not fundamentally address the inherent cost of multiplication in neural networks.
Consequently, multiplication-free architectures such as binary neural networks (BNNs) \cite{bnn}, AddNets \cite{addernet}, and DeepShift \cite{deepshift} have also been proposed, demonstrating impressive computational efficiency \cite{bnnapp1, bnnapp3, bnnapp4}.

\begin{figure}[t]
\centerline{\includegraphics[width=1.0\columnwidth]{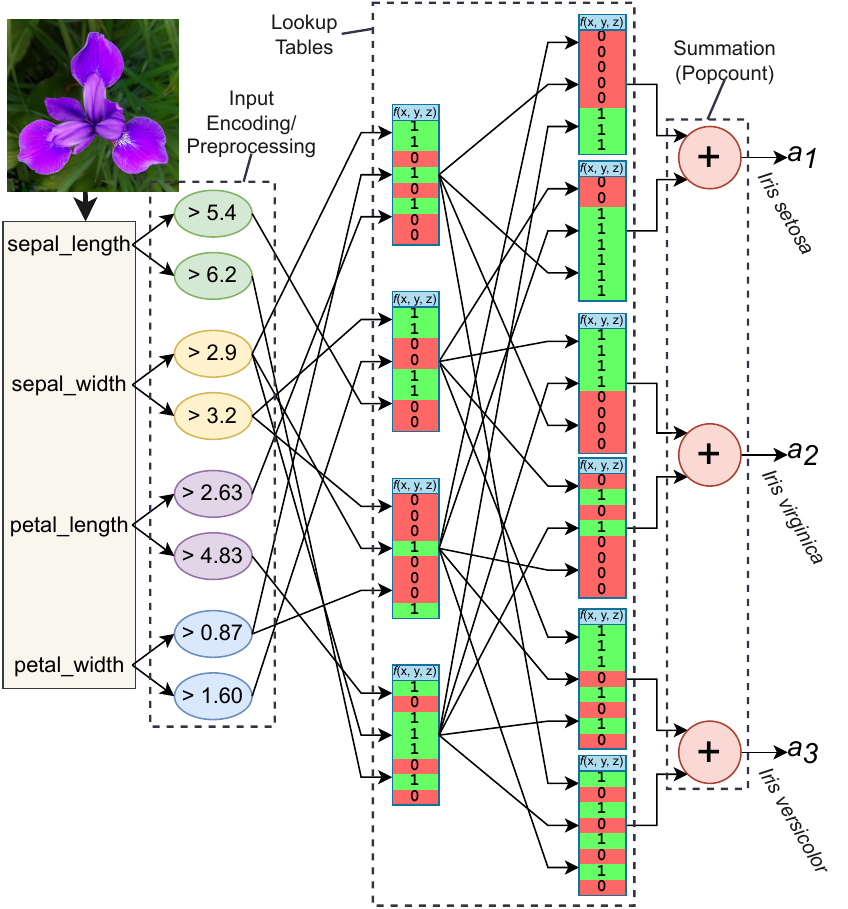}}
\caption{A very simple DWN for the Iris~\cite{misc_iris_53} dataset, shown at inference time.
DWNs perform computation using multiple layers of directly chained lookup tables (LUT-3s, in this example).
Inputs are binarized using a unary ``thermometer'' encoding, formed into tuples, and concatenated to address the first layer of LUTs.
Binary LUT outputs are used to form addresses for subsequent layers.
Outputs from the final layer of LUTs are summed to derive activations for each output class.
No arithmetic operations are performed between layers of LUTs.
}
\label{fig:iris}
\vspace{-5mm}
\end{figure}

Within the domain of multiplication-free models, weightless neural networks (WNNs) stand out as a distinct category. Diverging from the norm, WNNs forgo traditional weighted connections, opting instead for lookup tables (LUTs) with binary values to drive neural activity \cite{wisard, wnn_intro_esann}, where number of inputs to each LUT, \(n\), is a hyperparameter. 
This enables WNNs to represent highly nonlinear behaviors with minimal arithmetic.
However, a notable limitation of WNNs architectures is their restriction to single-layer models. This constraint is primarily due to the discrete structure of LUTs, which has historically 
made training complex multi-layer WNNs infeasible.



Despite this limitation, recent advancements have illuminated the potential of WNNs to achieve high efficiency in neural computation. The ULEEN model \cite{uleen} exemplifies this, showcasing a WNN's capabilities on an FPGA platform. Remarkably, ULEEN has outperformed BNNs in accuracy, energy efficiency, latency, memory consumption, and circuit area. This is particularly noteworthy considering that ULEEN operates with a single-layer structure, in contrast to the deep architecture of BNNs. This success not only underscores the inherent efficiency of WNNs but also implies the immense potential they could unlock if developed beyond their current single-layer constraints.

The recent Differentiable Logic Gate Networks (DiffLogicNet) \cite{difflogicnet} proposes a gradient descent-based method for training \textit{multi-layer} logic gate networks.
The hypothesis space of DiffLogicNet's two-input binary nodes is exactly the same as a WNN with two-input LUTs (LUT-2s), showing that training multi-layer WNNs is theoretically possible.
However, the cost of this method scales double-exponentially ($\mathcal{O}(2^{2^n})$) with the number of inputs to each LUT, requiring an astonishing \textit{18.4 quintillion} parameters to represent a single LUT-6.
The inability to train models with larger LUTs is detrimental for two main reasons: (1) the VC dimension of WNNs grows exponentially with the number of inputs to each LUT~\cite{vc_wi}, making a WNN with fewer but larger LUTs a more capable learner than one with many small LUTs; (2) the ability to train LUTs with varying sizes facilitates hardware-software co-design, leading to more efficient models. For instance, a LUT-6-based WNN could be significantly more efficient on modern AMD/Xilinx FPGAs, which employ LUT-6s in their configurable logic blocks (CLBs), aligning WNN implementation more closely with FPGA architecture.

Furthermore, both WNNs and DiffLogicNet currently have three other vital limitations. First, they rely on pseudo-random connections between LUTs, which leaves the optimal arrangement of the neural network to chance.
Second, they use population counts (popcounts) to determine activation values for each class, which incurs a large area overhead in hardware, with the popcount circuit in some cases being as large as the network itself.
Finally, the binary nature of LUTs means that traditional DNN regularization techniques are ineffective; thus, there is a pressing need to develop specialized regularization techniques.

Recognizing the critical need for innovation in this area, our work introduces the \textbf{\underline{D}ifferentiable \underline{W}eightless Neural \underline{N}etwork (DWN)}
(Figure \ref{fig:iris}), tackling all of these limitations.
This is achieved with a suite of innovative techniques:

\vspace{-3mm}
\begin{itemize}[leftmargin=*]
\setlength\itemsep{-3pt}
\item \textbf{Extended Finite Difference:}
 This technique enables efficient backpropagation through LUTs by approximate differentiation, allowing the development of multi-layer WNN architectures with bigger LUTs.
\item \textbf{Learnable Mapping:} This novel layer allows DWNs to learn the connections between LUTs during training, moving beyond fixed or random setups in WNNs and DiffLogicNet, enhancing adaptability and efficiency without additional overhead during inference.
\item \textbf{Learnable Reduction:} Tailored for tiny models, this approach replaces popcount with decreasing pyramidal LUT layers, leading to smaller circuit sizes.
\item \textbf{Spectral Normalization:} A normalization technique specifically developed for LUTs in WNNs to improve model stability and avoid overfitting.
\end{itemize}
\vspace{-2mm}
Our results demonstrate the DWN's versatility and efficacy in various scenarios: 
\vspace{-3mm}
\begin{enumerate}[leftmargin=*]
\setlength\itemsep{-3pt}
\item \textbf{FPGA deployment}: DWNs outperform DiffLogicNet, fully-connected BNNs, and prior WNNs in latency, throughput, energy efficiency, and model area across all tested datasets. By aligning model LUTs with the native FPGA LUT size, DWNs achieve a geometric average 2522$\times$ improvement in energy-delay product versus the FINN BNN platform and a 63$\times$ improvement versus ULEEN, the current state-of-the-art for efficient WNNs.
\item \textbf{Constrained edge devices}: On a low-end microcontroller (the Elegoo Nano), our throughput-optimized implementations of DWNs achieve on average 1.2\% higher accuracy than XGBoost with a 15\% speedup.
Our accuracy-optimized implementations achieve 5.4\% improvement, at the cost of execution speed.
\item \textbf{Ultra-low-cost chips:} The DWN reduces circuit area by up to 42.8$\times$ compared to leading Tiny Classifier models~\cite{tinyclassifier}, and up to 310$\times$ compared to DiffLogicNet.
\item \textbf{Tabular data}: DWN surpasses state-of-the-art models such as XGBoost and TabNets, achieving an average rank of 2.5 compared to 3.4 and 3.6 respectively.
\end{enumerate}

%% file: 02_background.tex
\section{Background \& Related Work}

\subsection{Weightless Neural Networks}
Weightless neural networks (WNNs) eschew traditional weighted connections in favor of a multiplication-free approach, using binary-valued lookup tables (LUTs), or ``RAM nodes'', to dictate neuronal activity. The connections between the input and these LUTs are randomly initialized and remain static. The absence of multiply-accumulate (MAC) operations facilitates the deployment of high-throughput models, which is particularly advantageous in edge computing environments. A notable recent work in this domain is ULEEN\cite{uleen}, which enhanced WNNs by integrating gradient-descent training and utilizing straight-through estimators \cite{ste} akin to those employed in BNNs, and outperformed the Xilinx FINN~\cite{finn} platform for BNN inference in terms of latency, memory usage, and energy efficiency in an FPGA implementation.
A significant limitation of most WNN architectures is their confinement to single-layer models. While some prior works experimented with multi-layer weightless models~\cite{mpln,gsn}, they relied on labyrinthine backward search strategies which were impractical for all but very simple datasets, and did not use gradient-based methods for optimization.


\subsection{Thermometer Encoding}

The method of encoding real-valued inputs into binary form is a critical aspect of WNNs, as the relationship between bit flips in the encoded input and corresponding changes in actual values is essential for effective learning \cite{binary_encodings}. To address this, Thermometer Encoding was introduced \cite{thermometer}, which uses a set of ordered thresholds to create a unary code (see Appendix \ref{apx:thermometer}).
\subsection{DiffLogicNet}

DiffLogicNet \cite{difflogicnet} proposed an approach to learning multi-layer networks exclusively composed of binary logic. In this model, an input binary vector is processed through multiple layers of binary logic nodes. These nodes are randomly connected, ultimately leading to a final summation determining the output class score. For training these networks via gradient descent, DiffLogicNet proposes a method where binary values are relaxed into probabilities. This is achieved by considering all possible binary logic functions (as detailed in Appendix A, Table \ref{tab:difflogic}), assigning a weight to each, and then applying a softmax function to create a probability distribution over these logic functions. See Appendix \ref{apx:difflogicnet} for more details.


\subsection{Other LUT-Based Neural Networks}
Recently, other LUT-based neural networks such as LogicNets \cite{logicnets}, PolyLUT \cite{polylut}, and NeuraLUT \cite{neurallut} have been proposed to improve DNN efficiency, rediscovering \cite{felipe1997, nsp}. LogicNets suggest training sparse DNNs with binary activations and converting their neurons into LUTs by considering all possible input combinations. This aims to achieve efficient inference but fails to fully utilize the computational capacity of LUTs. An \textit{n}-input LUT has a known VC-dimension of $2^n$ \cite{vc_wi}, while a DNN neuron with \textit{n} inputs has a VC-dimension of $n+1$. Consequently, they effectively train a LUT with a reduced VC-dimension of $n+1$, leading to larger and less efficient models.

\vspace{-1mm}
PolyLUTs tries to address this limitation by utilizing feature mappings in the sparse neuron inputs to learn more complex patterns. The most recent NeuraLUTs fit multiple neurons and layers with skip connections that receive the same input into a LUT, rather than a single neuron. However, both approaches still fall short of fully exploiting LUT computational capabilities, as we will demonstrate in the experiments section.

\vspace{-1mm}
In contrast, our approach fully leverages the computational capabilities of LUTs by proposing a method to update and perform backpropagation with LUTs during the training phase, rather than merely using LUTs as a speedup mechanism for DNN neurons or layers.


%% file: 03_methodology.tex
\vspace{-2mm}
\section{Methodology}

\subsection{Extended Finite Difference}

DiffLogicNet introduced a technique for learning binary logic with gradient descent and backpropagation that is readily applicable to WNNs employing two-input RAM nodes, as LUT-2s inherently represent binary logic. 
However, this technique is impractical for even slightly larger RAM nodes due to its $\mathcal{O}(2^{2^n})$ space and time complexities for a single LUT-$n$.
Crucially, our approach reduces this to $\mathcal{O}(2^n)$, cutting the weights and computations needed to represent a LUT-6 from 18,446,744,073,709,551,616 to 64.

\textbf{Finite Difference (FD)}
is a powerful tool for approximating derivatives, especially for functions with binary vector inputs. This approach is centered on evaluating the impact of minor input alterations on the input, specifically flipping a single bit in the binary case. For a given function \( f: \{0, 1\}^n \rightarrow \mathbb{R}^m \), the FD \( \Delta f \) is computed as $\Delta f(x)_j = f(\leftindex^1_j x) - f(\leftindex^0_j x)$ where \( x \) is the binary input vector, and $\leftindex^1_j {x}$ and $\leftindex^0_j {x}$ represents the vector $x$ with its $j$-th bit set to $1$ and $0$, respectively. This formula shows how \( f \)'s output changes when flipping the \( j \)-th bit in \( x \), capturing the output's sensitivity to specific input bits.

The derivatives of a lookup table's addressing function can be approximated using FD. Consider $A: \mathbb{R}^{2^n} \times \{0, 1\}^{n} \rightarrow \mathbb{R}$ as the addressing function that retrieves values from a lookup table $U \in \mathbb{R}^{2^n}$ using address $a \in \{0, 1\}^n$. Define $\delta: \{0, 1\}^n \rightarrow \{1, \ldots, 2^n\}$ as the function converting a binary string to its integer representation $+1$. The partial derivatives of $A$ can be approximated by finite differences:
\vspace{-1mm}
\[\frac{\partial A}{\partial U_i}(U, a) = 
 \begin{cases}
             1, & \text{if } i = \delta(a)\\
             0,              & \text{otherwise}
    \end{cases},
\]
\vspace{-1mm}
\[\frac{\partial A}{\partial a_j}(U, a) =  A(U, \leftindex^1_j{a}) - A(U, \leftindex^0_j{a})
\]
\vspace{-3mm}

where $\leftindex^1_j {a}$ and $\leftindex^0_j {a}$ signifies the address $a$ with its $j$-th bit set to $1$ and $0$, respectively.

Using FD is our first proposed approach to "Differentiable" WNNs (DWN). However, while FD approximates the partial derivatives of a lookup table's addressing function, it only considers addresses within a Hamming distance of 1 from the targeted position. This limitation may hinder learning by ignoring optimal addressing positions beyond this proximity. For example, in a LUT-6 scenario, FD considers only 7 out of 64 positions, potentially neglecting more relevant ones.

To address this limitation, we introduce an \textbf{Extended Finite Difference (EFD)} method for more comprehensive derivative approximation. This technique considers variations in the addressed position relative to all possible positions, not just those one-bit apart:
\vspace{-1mm}
\[
\frac{\partial A}{\partial a_j}(U, a) =  \sum_{k \in \{0, 1\}^n} \frac{{(-1)}^{(1-k_j)}A(U, k)}{H(k, a, j) + 1} 
\]
where $H: \{0, 1\}^n \times \{0, 1\}^n \times \mathbb{N} \rightarrow \mathbb{N}$ calculates the Hamming distance between $k$ and $a$, excluding the $j$-th bit. This formula integrates contributions from all lookup table positions, weighted by their relative distance (in terms of Hamming distance) to the address in use, with an added term for numerical stability. EFD provides a more holistic view, potentially capturing address shifts to more distant positions that conventional FD might miss.

\subsection{Learnable Mapping}

WNNs and DiffLogicNet both rely on pseudo-random mappings to route inputs, to LUTs in the former and between binary logic nodes in the latter.
The specific choice of mapping can have a substantial impact on model accuracy, but is largely dependent on chance.
In response, we introduce a new method that learns these connections through gradient descent-based optimization, without additional computational overhead during inference. This involves a weight matrix \( W \in \mathbb{R}^{P \times Q} \) during training, where \( P \) is the input bit length or output bit count from the previous layer, and \( Q \) is the number of input connections in the next layer. Input selection for LUTs during the forward pass is based on the maximum weights in \( W \), determined by \( I(W, x)_i = x_{\text{argmax}(W[i, :])} \).

The backward pass involves calculating partial derivatives with respect to \( W \) and input \( x \). For \( W \), we use the product of the transformed input vector (\(2x - 1\)) and the backpropagated gradient matrix \( G \), where the transformation maps binary inputs to \(-1\) and \(1\). The derivative is \( \frac{\partial I}{\partial W} = ((2x - 1)^\top \cdot G) \). For input \( x \), the derivative is obtained by multiplying \( G \) with the transposed softmax of \( W \) over the first dimension, as \( \frac{\partial}{\partial x} = G \cdot \text{softmax}_{\text{dim}=0}(W)^\top \). These gradients allow the learnable mapping (Figure \ref{fig:lm_and_lr}) to iteratively refine the LUTs connections, optimizing DWN performance. During inference, the argmax of \( W \) remains constant since it is independent of the input. Consequently, the weight matrix \( W \) is discarded, and the LUTs' connections become fixed, meaning there is no overhead from this technique at inference time.

\subsection{Learnable Reduction}
\begin{figure}
    \centering
    \includegraphics[width=0.95\columnwidth]{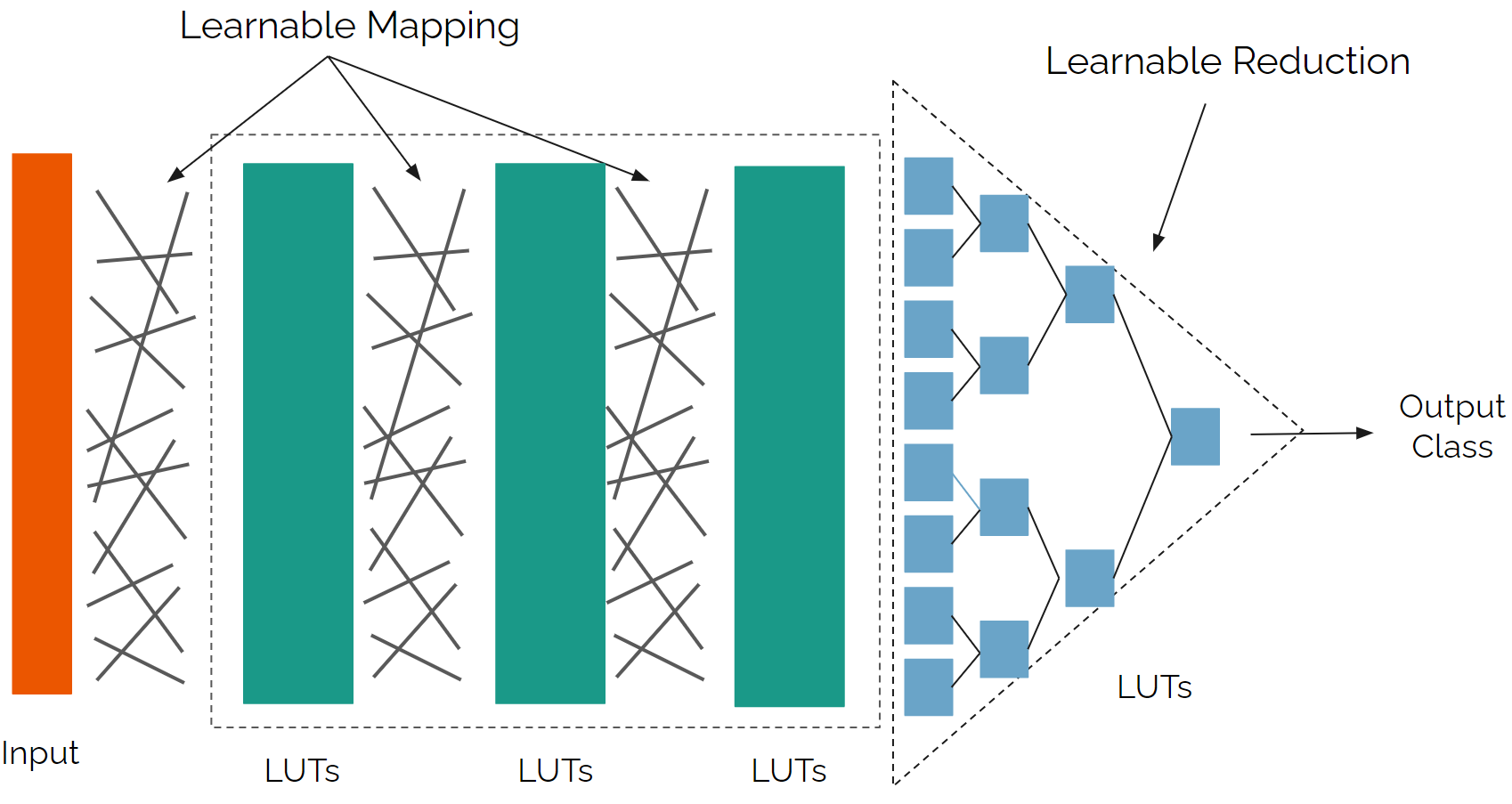}
    \vspace{-5mm}
    \caption{Learnable Mapping \& Learnable Reduction in DWNs.}
    \label{fig:lm_and_lr}
    \vspace{-4mm}
\end{figure}

When deploying tiny DWNs targetting ultra-low-cost chips, the popcount following the final LUT layer can constitute a large fraction of circuit area.
This size disparity hinders efforts to reduce the overall circuit size. To address this, we propose a novel approach that deviates from the conventional method of determining a WNN's output class; i.e., the argmax of popcounts from the last LUT layer's feature vector.
Instead, our method involves learning the reduction from the feature vector to the output class using layers of LUTs configured in a decreasing pyramidal architecture as in Figure \ref{fig:lm_and_lr}. This technique enables the model to discover more efficient methods for determining the output class. It moves away from the reliance on a fixed structure of popcounts and argmax computations, resulting in smaller and more efficient circuit designs suitable for deployment.

\subsection{Spectral Regularization}
The inherent nonlinearity of WNNs is a double-edged sword: it contributes to their remarkable efficiency but also makes them very vulnerable to overfitting.
Even very small DWNs may sometimes perfectly memorize their training data.
Unfortunately, conventional DNN regularization techniques can not be applied directly to DWNs.
For instance, since only the sign of a table entry is relevant for address computation, using L1 or L2 regularization to push entries towards 0 during training results in instability.

To address this issue, we propose a novel WNN-specific technique: \textit{spectral regularization}. For an $n$-input pseudo-Boolean function $f : \{0, 1\}^{n} \rightarrow \mathbb{R}$, we define the L2 spectral norm of $f$ as:
\[
    \frac{1}{2^{n}} \Bigl\lVert \Bigl\{ {\sum_{x \in \{0, 1\}^{n}}}\!\!f(x) \Bigl(\,\prod_{i \in S} (2x_{i} - 1)\Bigr) \;\Big\vert\; S \in [n] \Bigr\} \Bigr\rVert_{2}
\]
Note that this is simply the L2 norm of the Fourier coefficients of $f$~\cite{odonnell2021analysis}.
Additionally, since all terms except $f(x)$ are constant, we can precompute a coefficient matrix $\mathcal{C} \in \mathbb{R}^{2^{n} \times 2^{n}}$ to simplify evaluating the spectral norm at runtime.
In particular, for a layer of $u$ LUTs with $n$ inputs each and data matrix $L \in [-1,\,1]^{u \times 2^{n}}$, we express the spectral norm as: 
\vspace{-2mm}
\[
    \text{specnorm}(L) = \lVert L \mathcal{C} \rVert_{2},\;
    \mathcal{C}_{ij} := \frac{1}{2^{n}}\!\!\!\prod_{a \in \{b \;\vert\; i_{b} = 1\}} \!\!\!(2j_{a} - 1)
\]
The effect of spectral regularization is to increase the resiliency of the model to perturbations of single inputs.
For instance, if an entry in a RAM node is never accessed during training, but all locations at a Hamming distance of 1 away hold the same value, then the unaccessed location should most likely share this value.


%% file: 04_experiments.tex
\vspace{-2mm}
\section{Experimental Evaluation}
To demonstrate the effectiveness and versatility of DWNs, we evaluate them in several scenarios. First, we assess their performance on a custom hardware accelerator, implemented using a field-programmable gate array (FPGA), to demonstrate DWNs' extreme speed and energy efficiency in high-throughput edge computing applications.
Next, we implement DWNs on an inexpensive off-the-shelf microcontroller, demonstrating that they can operate effectively on very limited hardware, and emphasizing their practicality in cost-sensitive embedded devices.
We also consider the incorporation of DWNs into logic circuits, assessing their potential utility in ultra-low-cost chips.

Beyond hardware-focused evaluations, we also compare the accuracy of DWNs against state-of-the-art models for tabular data, with an emphasis on maximizing accuracy rather than minimizing model parameter size.
Overall, while DWNs are chiefly engineered for edge inference applications, we aim to demonstrate their effectiveness in multiple contexts.


\paragraph{Binary Encoding:} All datasets in the experimental evaluation are binarized using the Distributive Thermometer \cite{distributive} for both DWN and DiffLogicNet. The sole exception is the DiffLogicNet model for the MNIST dataset, for which we use a threshold of 0, following the strategy outlined in their paper.

\vspace{-.5mm}
\subsection{DWNs on FPGAs}
FPGAs enable the rapid prototyping of hardware accelerators without the lead times associated with fabricating a custom IC.
We deploy DWN models on the Xilinx Zynq Z-7045, an entry-level FPGA that was also used for the BNN-based FINN~\cite{finn} and WNN-based ULEEN~\cite{uleen}.
We adopt the input data compression scheme used in ULEEN, which allows for more efficient loading of thermometer-encoded values. This is important due to the limited (112 bits per cycle) interface bandwidth of this FPGA. As in prior work, all designs are implemented at a clock speed of 200 MHz.

\begin{figure}[htbp]
\vspace{-2mm}
\centerline{\includegraphics[width=0.95\columnwidth]{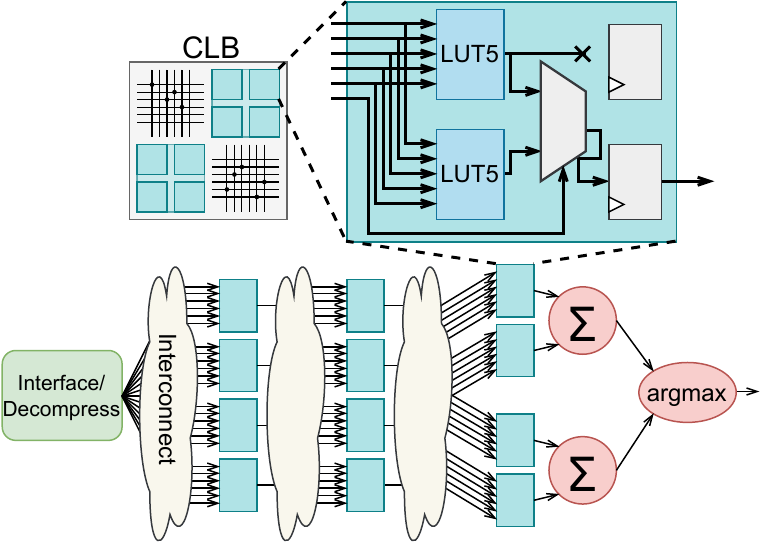}}
\vspace{-6mm}
\caption{Implementation of a DWN on an FPGA. Each hardware LUT-6 (subdivided into two LUT-5s and a 2:1 MUX) can implement a six-input RAM node. Registers buffer LUT outputs.}
\label{fig:fpga_layout}
\vspace{-3mm}
\end{figure}

Figure~\ref{fig:fpga_layout} gives a high-level overview of our accelerator design.
The FPGA is largely composed of configurable logic blocks (CLBs), which are in turn composed of six-input lookup tables (LUT-6s), flip-flops, and miscellaneous interconnect and muxing logic~\cite{xilinx_clb}.\footnote{As shown in Figure~\ref{fig:fpga_layout}, a LUT-6 can also function as two LUT-5s, but this requires both LUTs to have identical inputs, and is not something we explore in this paper.}
The Z-7045 provides a total of 218,600 LUT-6s, each of which can represent a six-input RAM node. Hence, DWNs with $n$=6 make efficient use of readily available FPGA resources.


\begin{table*}[ht]
\vspace{-2mm}
\centering
\setlength{\tabcolsep}{4pt}
\caption{Implementation results for DWNs and prior efficient inference models on a Z-7045 FPGA with input buffers and decompression implemented, with data being passed to the board using its I/O pins, which have a bandwidth of 112 bits per cycle, and the clock is limited to 200 MHz for a fair comparison to prior work. For results in out-of-context mode and with the clock is set to 250 MHz, see Appendix \ref{apx:ooc_dwn_fpga}. $^*$Model could not be synthesized; hardware values are approximate. $^\dag$ULEEN used augmented data for MNIST; we present MNIST results with and without augmentation.}

\label{tab:fpga_results}
\begin{tabular}{@{}llr@{\hspace{12pt}}r@{\hspace{12pt}}r@{\hspace{12pt}}r@{\hspace{12pt}}r@{\hspace{12pt}}r@{\hspace{4pt}}}
\toprule
\multicolumn{1}{c}{\multirow{2}{*}{\textbf{Dataset}}} & 
  \multicolumn{1}{c}{\multirow{2}{*}{\textbf{Model}}} &
  \multirow{2}{*}{\shortstack[c]{\textbf{Test}\\\textbf{Accuracy\%}}}\hspace{-5pt} &
  \multirow{2}{*}{\shortstack[c]{\textbf{Parameter}\\\textbf{Size (KiB)}}}\hspace{-5pt} &
  \multirow{2}{*}{\shortstack[c]{\textbf{Latency}\\\textbf{(ns)}}}\hspace{-5pt} &
  \multirow{2}{*}{\shortstack[c]{\textbf{Throughput}\\\textbf{(Samples/s)}}}\hspace{-5pt} &
  \multirow{2}{*}{\shortstack[c]{\textbf{Energy}\\\textbf{(nJ/Sample)}}}\hspace{-5pt} &
  \multirow{2}{*}{\shortstack[c]{\textbf{LUTs}\\\textbf{(1000s)}}}\hspace{-5pt} \\
& & & & & & & \\ \midrule
\multirow{8}{*}{MNIST}        & FINN                               & 98.40 & 355.3  & 2440 & 1.56M & 5445  & 83.0 \\
                              & ULEEN$^\dag$                          & 98.46 & 262.0 & 940  & 4.05M  & 823   & 123.1 \\
                              & DiffLogicNet \textit{(xs)}   & 96.87 & 11.7 & 90 & 33.3M & 17.2 & 9.6 \\
                              & DiffLogicNet \textit{(sm)}$^*$   & 97.62 & 23.4 & \estimated{95} & \estimated{33.3M} & --- & \estimated{19.1} \\
                              & DWN \textit{($n$=2; lg)}     & 98.27 & \textbf{5.9} & 135 & 25.0M & 42.3 & 10.3\\
                              & DWN \textit{($n$=6; sm)} & 97.80 & 11.7 & \textbf{60} & \textbf{50.0M} & \textbf{2.5} & \textbf{2.1} \\
                              & DWN \textit{($n$=6; lg)}   & 98.31 & 23.4  & 125  & 25.0M & 19.0  & 4.6 \\
                              & DWN \textit{($n$=6; lg; +aug)}$^\dag$ & \textbf{98.77} & 23.4  & 135  & 22.2M & 21.6  & 4.6 \\
                              \midrule
\multirow{5}{*}{FashionMNIST}
                              & FINN                        & 84.36 & 355.3  & 2440  & 1.56M   & 5445   & 83.0 \\
                              & ULEEN                       & 87.86 & 262.0 & 940  & 4.05M   & 823   & 123.1 \\
                              & DiffLogicNet    & 87.44 & 11.7 & 270 & 9.52M & 119.8 & 11.4 \\
                              & DWN \textit{($n$=2)} & \textbf{89.12} & \textbf{7.8} & 255 & \textbf{10.0M} & 145.4 & 13.6 \\
                              & DWN \textit{($n$=6)}  & 89.01 & 31.3  & \textbf{250} & \textbf{10.0M} & \textbf{90.9} & \textbf{7.6} \\
\midrule
\multirow{5}{*}{KWS} & FINN & 70.60 & 324 & 7780 & 0.67M & 5716 & 42.8 \\
                     & ULEEN & 70.34 & 101.0 & 390 & 10.0M & 642 & 141.1 \\
                     & DiffLogicNet$^*$ & 64.18 & 23.4 & \estimated{265} & \estimated{10.0M} & --- & \estimated{20.3} \\
                     & DWN \textit{($n$=2)} & 70.92 & \textbf{2.9} & \textbf{235} & \textbf{10.5M} & 79.2 & 6.8 \\
                     & DWN \textit{($n$=6)} & \textbf{71.52} & 12.5 & \textbf{235} & \textbf{10.5M} & \textbf{42.3} & \textbf{4.8} \\
\midrule
\multirow{6}{*}{ToyADMOS/car} & FINN & 86.10 & 36.1 & 3520 & 1.57M & 547 & 14.1 \\
                              & ULEEN & 86.33 & 16.6 & 340 & 11.1M & 143 & 29.4 \\
                              & DiffLogicNet & 86.66 & 3.1 & 165 & 16.7M & 23.2 & 3.9 \\
                              & DWN \textit{($n$=2; sm)} & 86.68 & \textbf{0.9} & \textbf{115} & \textbf{25.0M} & 7.6 & 2.2 \\
                              & DWN \textit{($n$=6; sm)} & 86.93 & 3.1 & 120 & 22.2M & \textbf{5.8} & \textbf{1.3} \\
                              & DWN \textit{($n$=6; lg)} & \textbf{89.03} & 28.1 & 165 & 16.7M & 45.4 & 6.2 \\
\midrule
\multirow{5}{*}{CIFAR-10} & FINN & \textbf{80.10} & 183.1 & 283000 & 21.9K & 150685 & 46.3 \\
                         & ULEEN & 54.21 & 1379 & --- & --- & --- & --- \\
                         & DiffLogicNet$^*$ & 57.29 & 250.0 & \estimated{11510} & \estimated{87.5K} & --- & \estimated{283.3} \\
                         & DWN \textit{($n$=2)}$^*$ & 57.51 & \textbf{23.4} & \estimated{2200} & \estimated{468K} & --- & \estimated{45.7} \\
                         & DWN \textit{($n$=6)} & 57.42 & 62.5 & \textbf{2190} & \textbf{468K} & \textbf{3972} & \textbf{16.7} \\
\bottomrule
\end{tabular}
\vspace{-3mm}
\end{table*}

\begin{table*}[]
\centering
\caption{FPGA implementation results comparing DWNs with other LUT-based neural networks. Results for LogicNets, PolyLUT, and NeuraLUT are sourced from their respective papers, while new LogicNets results ($^*$) are from their official GitHub~\cite{logicnets_github}. "Time/Sample" denotes the inference time per sample (calculated as 1/Throughput). Latency refers to the total end-to-end time of the model execution.}
\label{tab:lutmodels}
\setlength{\tabcolsep}{4pt}
\begin{tabular}{@{}llrrrrrrcrc@{}}
\toprule
  \multirowcell{2}{\bf Dataset} &
  \multirowcell{2}{\bf Model} &
  \multirowcell{2}{\bf Accuracy} &
  \multirowcell{2}{\bf LUT} &
  \multirowcell{2}{\bf FF} &
  \multirowcell{2}{\bf DSP} &
  \multirowcell{2}{\bf BRAM} &
  \multirowcell{2}{\bf Fmax \\\bf (MHz)} &
  \multirowcell{2}{\bf Time/Sample} &
  \multirowcell{2}{\bf Latency} &
  \multirowcell{2}{\bf Area$\bm \times$Lat. \\\bf (LUT×ns)} \\\\ \midrule
\multirow{6}{*}{MNIST} &
  PolyLUT &
  96\% &
  70673 &
  4681 &
  0 &
  0 &
  378 &
  2.6ns &
  16.0ns &
  1.1e+06 \\
  &
  NeuraLUT &
  96\% &
  54798 &
  3757 &
  0 &
  0 &
  431 &
  2.3ns &
  12.0ns &
  6.6e+05 \\
 &
  DWN \textit{($n$=6; sm)} &
  97.1\% &
  \textbf{692} &
  \textbf{422} &
  0 &
  0 &
  827 &
  1.2ns &
  \textbf{2.4ns} &
  \textbf{1.6e+03} \\
  &
  DWN \textit{($n$=6; md)} &
  97.8\% &
  2055 &
  1675 &
  0 &
  0 &
  \textbf{873} &
  \textbf{1.1ns} &
  4.6ns &
  9.4e+03 \\
  &
  DWN \textit{($n$=6; md)} &
  97.9\% &
  1413 &
  1143 &
  0 &
  0 &
  827 &
  1.2ns &
  3.6ns &
  5.1e+03 \\
  &
  DWN \textit{($n$=6; lg)} &
  \textbf{98.3\%} &
  4082 &
  3385 &
  0 &
  0 &
  827 &
  1.2ns &
  6.0ns &
  2.4e+04 \\
\midrule
\multirow{7}{*}{JSC} &
  LogicNets &
  71.8\% &
  37931 &
  810 &
  0 &
  0 &
  427 &
  2.3ns &
  13.0ns &
  4.9e+05 \\
  &
  LogicNets \textit{(sm)}* &
  69.8\% &
  244 &
  270 &
  0 &
  0 &
  1353 &
  0.7ns &
  5.0ns &
  1.2e+03 \\
  &
  LogicNets \textit{(lg)}* &
  73.1\% &
  36415 &
  2790 &
  0 &
  0 &
  390 &
  2.6ns &
  6.0ns &
  2.2e+05 \\
  &
  PolyLUT &
  72\% &
  12436 &
  773 &
  0 &
  0 &
  646 &
  1.5ns &
  5.0ns &
  6.2e+04 \\
 &
  NeuraLUT &
  72\% &
  4684 &
  341 &
  0 &
  0 &
  727 &
  1.4ns &
  3.0ns &
  1.4e+04 \\
  &
  DWN \textit{($n$=6; sm)} &
  71.1\% &
  \textbf{20} &
  \textbf{22} &
  0 &
  0 &
  \textbf{3030} &
  \textbf{0.3ns} &
  \textbf{0.6ns} &
  \textbf{1.3e+01} \\
  &
  DWN \textit{($n$=6; sm)} &
  \textbf{74.0\%} &
  110 &
  72 &
  0 &
  0 &
  1094 &
  0.9ns &
  1.5ns &
  2.0e+02 \\ 
  \midrule
\multirow{3}{*}{JSC} &
  PolyLUT &
  75\% &
  236541 &
  2775 &
  0 &
  0 &
  235 &
  4.3ns &
  21.0ns &
  5.0e+06 \\ 
  &
  NeuraLUT &
  75\% &
  92357 &
  4885 &
  0 &
  0 &
  368 &
  2.7ns &
  14.0ns &
  1.3e+06 \\
  &
  DWN \textit{($n$=6; md)} &
  \textbf{75.6\%} &
  \textbf{720} &
  \textbf{457} &
  0 &
  0 &
  \textbf{827} &
  \textbf{1.2ns} &
  \textbf{3.6ns} &
  \textbf{2.6e+03} \\
  \midrule
\multirow{2}{*}{JSC} &
  hls4ml &
  76.2\% &
  63251 &
  4394 &
  38 &
  0 &
  200 &
  - &
  45.0ns &
  2.8e+06 \\
 &
  DWN \textit{($n$=6; lg)} &
  \textbf{76.3\%} &
  \textbf{4972} &
  \textbf{3305} &
  0 &
  0 &
  \textbf{827} &
  \textbf{1.2ns} &
  \textbf{7.3ns} &
  \textbf{3.6e+04} \\
\bottomrule
\end{tabular}
\vspace{-10pt}
\end{table*}

Table~\ref{tab:fpga_results} compares the FPGA implementations of our DWN models against prior work.
We include DWNs with both two and six-input RAM nodes.
The original DiffLogicNet paper~\cite{difflogicnet} does not propose an FPGA implementation, but we observe that their model is structurally identical at inference time to DWNs with $n$=2 inputs per LUT, with all substantive differences restricted to the training process.
Therefore, we can directly implement their models using our DWN RTL flow.

All datasets were chosen due to their use in prior work except for FashionMNIST~\cite{xiao2017/online}, which is identical in size to MNIST but intentionally more difficult. We directly reuse the MNIST model topologies, and thus hardware results, for FINN and ULEEN on this dataset.

Excluding CIFAR-10, our DWN models are smaller, faster, and more energy-efficient than prior work, with comparable or better accuracy.
In particular, latency, throughput, energy per sample, and hardware area (in terms of FPGA LUTs) are improved by geometric averages of
\mbox{(20.7, 12.3, 121.6, 11.7)$\times$} respectively versus FINN, and \mbox{(3.3, 2.3, 19.0, 22.7)$\times$} respectively versus ULEEN,
the prior state-of-the-art for efficient WNNs. Unlike the other architectures in Table~\ref{tab:fpga_results}, FINN supports convolution. This gives it vastly superior accuracy on the CIFAR-10 dataset, albeit at a hefty penalty to speed and energy efficiency.

Several models in Table~\ref{tab:fpga_results} could not be implemented on our target FPGA (indicated by `$*$').
The primary cause of this was routing congestion: since it would be infeasibly expensive for FPGAs to implement a full crossbar interconnect, they instead have a finite number of wires to which they assign signals during synthesis.
The irregular connectivity between layers in DiffLogicNet and DWNs with $n$=2 proved impossible to map to the FPGA's programmable interconnect in these cases.
However, note that all DWNs with $n$=6 were successfully routed and implemented.

An interesting takeaway from these results is that the parameter sizes of DWN models are not necessarily good predictors of their hardware efficiency. For instance, the large MNIST model with $n$=2 has $\approx$1/4 the parameter size of the $n$=6 model, yet more than twice the area and energy consumption.
Since our target FPGA uses LUT-6s natively, models with $n$=6 are inherently more efficient to implement. Although the synthesis tool can perform logic optimizations that map multiple DWN LUT-2s to a single FPGA LUT-6, this is not enough to offset the $\approx$4$\times$ larger number of RAM nodes needed to achieve the same accuracy with $n$=2.


\paragraph{Comparison to Other LUT-Based NNs:} We also compare DWNs against LogicNets, PolyLUT, and NeuraLUT, which convert models to LUTs for inference but do not use them during training.
We follow their experimental methodology by targeting the \texttt{xcvu9p\-flgb2104-2-i} FPGA, running synthesis with Vivado's \texttt{Flow\_PerfOptimized\_high} strategy in out-of-context mode, and using the highest possible clock frequency.
We use the MNIST and Jet Substructure (JSC) datasets, as in the NeuraLUT paper, and compare our models against published results. This comparison is shown in Table~\ref{tab:lutmodels}.


DWNs achieves superior accuracy with significantly reduced LUT usage and area-delay product (ADP) compared to all other methods. Notably, on JSC ($ \geq $75\%), our \textit{md} model achieves slightly higher accuracy then NeuraLUT with 43.1$\times$ fewer LUTs, 1.57$\times$ reduced latency, and 67.8$\times$ reduced ADP.


\subsection{DWNs on Microcontrollers}


While FPGAs can be extraordinarily fast and efficient, they are also expensive, specialized devices.
We also consider the opposite extreme: low-cost commodity microcontrollers.
The Elegoo Nano is a clone of the open-source Arduino Nano, built on the 8-bit ATmega328P, which at the time of writing retails for \$1.52 in volume.
The ATmega provides 2 KB of SRAM and 30 KB of flash memory and operates at a maximum frequency of 20 MHz.
We can not expect performance comparable to an FPGA on such a limited platform.
Our goal is instead to explore the speeds and accuracies of DWNs which can fit into this device's memory.

We use two strategies for implementing DWNs on the Nano. Our first approach uses aggressive bit packing to minimize memory usage, allowing us to fit more complex models on the device. For instance, the 64 entries of a LUT-6 can be packed in 8 bytes of memory, and the six indices for its inputs can be stored in 7.5 bytes by using 10-bit addresses (for more details on our bit-packing strategy, see Appendix~\ref{apx:microcontroller}).
However, this approach needs to perform bit manipulation to unpack data, which is fairly slow.
Therefore, we also explore an implementation without bit-packing, which greatly increases inference speed but reduces the maximum size (and therefore accuracy) of feasible models.

XGBoost~\cite{xgboost} is a widely-used tree boosting system notable for its ability to achieve high accuracies with tiny parameter sizes.
This makes it a natural choice for deployment on microcontrollers.
We use the MicroML~\cite{micromlgen} library for XGBoost inference on the Nano and compare it against DWNs.
To fit entire samples into SRAM, we quantize inputs to 8 bits. We did not observe a significant impact on accuracy from this transformation.

Table~\ref{tab:arduino} compares DWNs against XGBoost on the Nano. We present results for the datasets from Table~\ref{tab:fpga_results}, excluding CIFAR-10, which was too large to fit even after quantization.
We also include three additional \textit{tabular} datasets, a category on which XGBoost excels.
All XGBoost models have a maximum tree depth of 3, with forest size maximized to fill the board's memory. We found that this gave better results than the default max depth of 6, which required extremely narrow forests in order to fit.
The parameter sizes of these XGBoost models are generally quite small, but their complex control flow means that their source code footprints are large, even after compiler optimizations (with \texttt{-Os}).

\begin{table}[ht]
\vspace{-2mm}
  \centering
  \setlength{\tabcolsep}{2.7pt}
  \caption{Model accuracies and throughputs (in inferences per second) for DWNs and XGBoost on the Elegoo Nano, a low-end microcontroller. All models are as large as possible within the constraints of the device's memory. We consider an accuracy-optimized DWN implementation which uses bit-packing, and a throughput-optimized implementation which does not.}
  \begin{tabular}{lrrrrrr}
    \toprule
\multicolumn{1}{c}{\multirow{3}{*}{\textbf{Dataset}}} &
    \multicolumn{4}{c}{\textbf{DWN}} &
    \multicolumn{2}{c}{\multirow{2}{*}{\textbf{XGBoost}}} \\
    \cmidrule(lr){2-5}
& \multicolumn{2}{c}{\textbf{Acc-Optim}} & \multicolumn{2}{c}{\textbf{Thrpt-Optim}} & & \\
    \cmidrule(lr){2-3} \cmidrule(lr){4-5} \cmidrule(lr){6-7}
& \multicolumn{1}{c}{\textbf{Acc.}} & \multicolumn{1}{c}{\textbf{Thrpt}} & \multicolumn{1}{c}{\textbf{Acc.}} & \multicolumn{1}{c}{\textbf{Thrpt}} & \multicolumn{1}{c}{\textbf{Acc.}} & \multicolumn{1}{c}{\textbf{Thrpt}} \\
    \midrule
    
    MNIST & 97.9\%\hspace{-4pt} & 16.5\scriptsize{/s} & 94.5\%\hspace{-4pt} & 108\scriptsize{/s} & 90.2\%\hspace{-4pt} & 81\scriptsize{/s} \\
    F-MNIST & 88.2\%\hspace{-4pt} & 16.4\scriptsize{/s} & 84.1\%\hspace{-4pt} & 95\scriptsize{/s} & 83.2\%\hspace{-4pt} & 81\scriptsize{/s} \\
    KWS & 69.6\%\hspace{-4pt} & 16.4\scriptsize{/s} & 53.6\%\hspace{-4pt} & 109\scriptsize{/s} & 51.0\%\hspace{-4pt} & 103\scriptsize{/s} \\
    ToyADMOS & 88.7\%\hspace{-4pt} & 17.7\scriptsize{/s} & 86.1\%\hspace{-4pt} & 112\scriptsize{/s} & 85.9\%\hspace{-4pt} & 94\scriptsize{/s} \\
    phoneme & 89.5\%\hspace{-4pt} & 17.8\scriptsize{/s} & 87.5\%\hspace{-4pt} & 298\scriptsize{/s} & 86.5\%\hspace{-4pt} & 265\scriptsize{/s} \\
    skin-seg & 99.8\%\hspace{-4pt} & 17.5\scriptsize{/s} & 99.7\%\hspace{-4pt} & 298\scriptsize{/s} & 99.4\%\hspace{-4pt} & 268\scriptsize{/s} \\
    higgs & 72.4\%\hspace{-4pt} & 17.1\scriptsize{/s} & 71.2\%\hspace{-4pt} & 254\scriptsize{/s} & 71.8\%\hspace{-4pt} & 245\scriptsize{/s} \\
    \bottomrule
  \end{tabular}
  \label{tab:arduino}
  \vspace{-2pt}
\end{table}


Our bit-packed DWN implementation is consistently more accurate than XGBoost, by an average of 5.4\%, and particularly excels on non-tabular multi-class datasets such as MNIST and KWS.
However, it is also 8.3$\times$ slower on average.
Our unpacked implementation is 15\% faster than XGBoost and still 1.2\% more accurate on average, but is less accurate on one dataset (higgs).
Overall, DWNs are good models for low-end microcontrollers when accuracy is the most important consideration, but may not always be the best option when high throughput is also needed.

\subsection{DWNs for Ultra-Low-Cost Chips}

To assess the viability of DWNs for ultra-low-cost chip implementations, we analyze their performance in terms of accuracy and NAND2 equivalent circuit area, comparing them with Tiny Classifiers~\cite{tinyclassifier_nature}, a SOTA work for ultra-low-cost small models, and DiffLogicNet. The datasets for this analysis are those shared between the Tiny Classifiers  (see Appendix~\ref{apx:datasets}) and AutoGluon (to be used in the next subsection) studies, providing a consistent basis for comparison. We also adhere to their data-splitting methods, using 80\% of the data for the training set and 20\% for the testing set.

Our DWN, utilizing the Learnable Reduction technique with LUT-2s, is designed to inherently learn two input binary logic, which directly correlates to logic gate formation in a logic circuit. The NAND2 equivalent size of our model is calculated by converting each LUT-2 into its NAND2 equivalent (e.g., a LUT-2 representing an OR operation equates to 3 NAND gates). For DiffLogicNet, we adopt a similar approach, translating the converged binary logic nodes into their NAND2 equivalents, plus the additional NAND2 equivalent size required for each class output popcount (Appendix~\ref{apx:popcount}), as per their model architecture. Notably, our DWN model, due to Learnable Reduction, does not incur this additional computational cost. For Tiny Classifiers we utilize the results reported in their paper.

Our results, presented in Table~\ref{tab:tiny_tabular}, highlight DWN's exceptional balance of efficiency and accuracy across a range of datasets. Notably, DWN consistently outperforms Tiny Classifiers and DiffLogicNet in accuracy, while also showcasing a remarkable reduction in model size. In the `skin-seg' dataset, DWN achieves 98.9\% accuracy with a model size of only 88 NAND, compared to 93\% with 300 NAND for Tiny Classifiers and 98.2\% with 610 NAND for DiffLogicNet, demonstrating reductions of approximately 3.4$\times$ and 6.9$\times$, respectively. Similarly, in the `jasmine' dataset, DWN reaches 80.6\% accuracy with just 51 NAND gates, while Tiny Classifiers and DiffLogicNet achieve 72\% with 300 NAND and 76.7\% with 1816 NAND, respectively, indicating reductions of 5.9$\times$ and 35.6$\times$. 

These findings demonstrate DWN's potential in ASIC and Ultra-Low-Cost Chip implementations, offering a blend of high accuracy and compact circuit design.

\begin{table}[htbp]
\vspace{-4mm}
\centering
\caption{Comparison of the accuracy and NAND2 equivalent circuit size for DWN, Tiny Classifiers, and DiffLogicNet across various datasets. This comparison highlights DWN's increased accuracy and significantly smaller circuit sizes, underscoring its effectiveness for ultra-low-cost chip implementations.}
\label{tab:tiny_tabular}
\setlength{\tabcolsep}{2.5pt}
\begin{tabular}{@{}llrlrlr@{}}
\toprule
\multirow{2}{*}{\textbf{Dataset}\hspace{15pt}} & \multicolumn{2}{c}{\textbf{DWN}} & \multicolumn{2}{c}{\textbf{DiffLogicNet}} & \multicolumn{2}{c}{\textbf{Tiny Class.}} \\
\cmidrule(lr){2-3}  \cmidrule(lr){4-5}  \cmidrule(lr){6-7}
           & \textbf{Acc.}   & \textbf{NAND} & \textbf{Acc.}   & \textbf{NAND}  & \textbf{Acc.} & \textbf{NAND} \\ \midrule
phoneme    & \textbf{85.7}\% & \textbf{168}\hspace{4pt}  & 83.2\% & 836\hspace{4pt}  & 79\% & 300\hspace{4pt}    \\
skin-seg   & \textbf{98.9}\% & \textbf{88}\hspace{4pt}   & 98.2\% & 610\hspace{4pt}   & 93\% & 300\hspace{4pt}    \\
higgs      & \textbf{67.8}\% & \textbf{94}\hspace{4pt}   & 67.5\% & 658\hspace{4pt}   & 62\% & 300\hspace{4pt}    \\
australian & \textbf{88.5}\% & \textbf{7}\hspace{4pt}    & 87.7\% & 379\hspace{4pt}   & 85\% & 300\hspace{4pt}    \\
nomao      & \textbf{93.5}\% & \textbf{87}\hspace{4pt}   & 93.5\% & 4955\hspace{4pt}  & 80\% & 300\hspace{4pt}    \\
segment    & \textbf{99.4}\% & \textbf{71}\hspace{4pt}   & 99.4\% & 610\hspace{4pt}   & 95\% & 300\hspace{4pt}    \\
miniboone  & \textbf{90.1}\% & \textbf{60}\hspace{4pt}   & 90.1\% & 619\hspace{4pt}  & 82\% & 300\hspace{4pt}    \\
christine  & \textbf{70.6}\% & \textbf{53}\hspace{4pt}   & 68.3\% & 16432\hspace{4pt} & 59\% & 300\hspace{4pt}    \\
jasmine    & \textbf{80.6}\% & \textbf{51}\hspace{4pt}   & 76.7\% & 1816\hspace{4pt}  & 72\% & 300\hspace{4pt}    \\
sylvine    & \textbf{92.2}\% & \textbf{44}\hspace{4pt}   & 85.7\% & 501\hspace{4pt}   & 89\% & 300\hspace{4pt}    \\
blood      & \textbf{78.4}\% & \textbf{49}\hspace{4pt}   & 78.4\% & 365\hspace{4pt}   & 63\% & 300\hspace{4pt}   \\
\bottomrule
\end{tabular}
\vspace{-15pt}
\end{table}

\begin{table*}[t!]
\vspace{-2mm}
\centering
\caption{A comprehensive evaluation of the accuracy of our proposed DWN against prominent state-of-the-art models in handling tabular data. Key metrics include the average ranking (Avg Rank), indicating each model's relative rank across datasets, and the average L1 norm (Avg Dist) from the top accuracy per dataset, assessing how closely each model approaches the best performance, with lower values indicating superior performance for both metrics.}
\label{tab:big_tabular}
\setlength{\tabcolsep}{4.7pt}
\begin{tabular}{@{}lcccccccc@{}}
\toprule
\multicolumn{1}{c}{\multirow{2}{*}{\textbf{Dataset}}} &
  \multirow{2}{*}{\textbf{DWN}} &
  \multirow{2}{*}{\textbf{DiffLogicNet}} &
  \textbf{AutoGluon} &
  \textbf{AutoGluon} &
  \textbf{AutoGluon} &
  \textbf{AutoGluon} &
  \textbf{AutoGluon} &
  \textbf{Google} \\
\multicolumn{1}{c}{} &
   &
   &
  \textbf{XGBoost} &
  \textbf{CatBoost} &
  \textbf{LightGBM} &
  \textbf{TabNN} &
  \textbf{NNFastAITab} &
  \textbf{TabNet} \\ \midrule
  
phoneme           & \textbf{0.895} & 0.891          & 0.886          & 0.868          & 0.873          & 0.884 & \textbf{0.895} & 0.844 \\
skin-seg          & \textbf{1.000}  & 0.999          & \textbf{1.000}  & \textbf{1.000}  & 0.999          & 0.999 & \textbf{1.000}  & 0.999 \\
higgs             & 0.727          & 0.711          & 0.728          & 0.730  & \textbf{0.743} & 0.731 & 0.727          & 0.726 \\
australian        & \textbf{0.901} & 0.862          & 0.870            & 0.862          & 0.870            & 0.870            & 0.855          & 0.529          \\
nomao             & 0.966          & 0.966          & \textbf{0.973}          & 0.963          & 0.964          & 0.972 & \textbf{0.973} & 0.959 \\
segment           & \textbf{0.998} & \textbf{0.998} & 0.989 & \textbf{0.998} & 0.996          & 0.996          & 0.994          & 0.857          \\
miniboone         & 0.946          & 0.944          & 0.948 & \textbf{0.952} & 0.860   & 0.948 & 0.947 & 0.717          \\
christine         & 0.736          & 0.710            & 0.750            & 0.728          & 0.734          & 0.754 & \textbf{0.756} & 0.547 \\
jasmine           & \textbf{0.816} & \textbf{0.816} & 0.812          & \textbf{0.816}          & 0.781          & 0.806          & 0.809          & 0.759          \\
sylvine           & 0.952          & 0.945          & 0.944          & 0.921          & 0.941          & 0.951 & \textbf{0.953} & 0.921 \\
blood             & 0.780            & 0.760            & 0.773 & \textbf{0.787} & \textbf{0.787} & 0.753 & 0.753          & 0.740   \\ \midrule
\textbf{Avg Rank} & \textbf{2.5}  & 4.5           & 3.4          & 3.6           & 4.5           & 3.6           & 3.5           & 7.5           \\
\textbf{Avg L1} & \textbf{0.005}  & 0.016           & 0.009           & 0.014           & 0.021           & 0.010           & 0.010          & 0.107       \\
\bottomrule
\end{tabular}
\vspace{-2mm}
\end{table*}

\subsection{DWNs on Tabular Data} 
In this subsection, we explore benchmarking DWNs against a range of prominent state-of-the-art models in the field of tabular data processing. This includes a thorough evaluation alongside the AutoGluon suite \cite{autogluon} — encompassing models like XGBoost, CatBoost, LightGBM, TabNN, and NNFastAITab — Google's TabNet, and DiffLogicNet. These benchmarks are crucial in demonstrating the efficacy and competitiveness of DWN in handling structured data, a key requirement for numerous real-world applications. Additionally, they show that DWN's efficient inference does not come at the cost of accuracy, highlighting DWN's remarkable ability to learn from tabular data.

To ensure a fair comparison, all models in the AutoGluon suite were trained under 'best quality' configurations, which involve extended training times and fine-tuning for optimal accuracy.
DWN model sizes were restricted to match those of the other models, ensuring that any performance gains were not simply due to a larger model size.  The datasets and train-test splits are the same as in the previous subsection. 

For training our DWN, we adopted an AutoML-like approach: setting the number of layers to 3 for small datasets and 5 for larger ones. The number of LUTs per layer was adjusted to align with the final model sizes of XGBoost, thereby maintaining a comparable model size. See Appendix \ref{apx:modelconfig} for more model configuration details.

Results are detailed in Table \ref{tab:big_tabular}. Key metrics in our analysis are the Average Rank and Average L1 norm. Average Rank is calculated by ranking models on each dataset according to accuracy and then averaging these ranks across all datasets. This metric provides a comparative view of each model's performance relative to others. The Average L1 norm, on the other hand, measures the average L1 distance of each model's accuracy from the highest-achieving model on each dataset. This offers insight into how closely each model approaches the best possible accuracy.

As shown in Table~\ref{tab:big_tabular}, DWN achieves an impressive average rank of 2.5 and an average L1 norm of 0.005, indicating its leading performance in accuracy among the compared models. Notably, it surpasses renowned models such as XGBoost, CatBoost, and TabNN, which have respective average rankings of 3.4, 3.6, and 3.6, and average L1 norms of 0.009, 0.014, and 0.010.


%% file: 05_conclusion.tex
\vspace{-3mm}
\section{Conclusion}
In this paper, we have introduced the Differentiable Weightless Neural Network (DWN), which features the novel Extended Finite Difference technique along with three significant enhancements: Learnable Mapping, Learnable Reduction, and Spectral Regularization. Our results underscore the versatility and efficiency of our approach, demonstrating up to 135$\times$ reduction in energy costs in FPGA implementations compared to BNNs and DiffLogicNet, up to 9\% higher accuracy in deployments on constrained devices, and culminating in up to 42.8$\times$ reduction in circuit area for ultra-low-cost chip implementations. Moreover, DWNs have achieved an impressive average ranking of 2.5 in processing tabular datasets, outperforming state-of-the-art models such as XGBoost and TabNets, which have average rankings of 3.4 and 3.6 respectively.

These significant contributions pave the way for a plethora of future work. In the context of FPGAs, extending our approach to include CNN and Transformer architectures could lead to significant advancements in deep learning. This is especially relevant considering the resource-intensive nature of current models used in computer vision and language processing. Furthermore, immediate future work on FPGAs could involve applying our models to domains where high-throughput or low latency are critical, such as anomaly detection, data streaming, and control systems. In the realm of ultra-low-cost chip implementations, DWNs hold significant promise for smart applications such as heart monitoring and seizure detection. These chips necessitate extremely compact architectures and operation on very low power, making them ideal for our approach. Furthermore, in the field of tabular data, the application of Differentiable Weightless Neural Networks (DWNs) could be transformative. Areas such as predictive analytics, financial modeling, and healthcare data analysis could greatly benefit from the efficiency and accuracy of DWNs, particularly in managing large and complex datasets.
Overall, DWNs show great potential for a range of edge and tabular applications.

%% file: A0_appendix.tex
\newpage
\appendix
\onecolumn
\section{Table of Boolean Functions}



\begin{table}[ht]
    \centering
    \caption{There are a total of $2^{2^n}$ logic functions for n logic variables. This table shows the 16 different functions for 2 variables. For 3 variables, there will be 256 logic operators. This list shows the functions in logic function and real-valued forms. The four columns on the right show the 4 values for each function for the input variable values of 00, 01, 10 and 11.  
    }
    \label{tab:difflogic}
    \begin{tabular}{cccccccc}
        \toprule
        ID & Operator & Real-valued & 00 & 01 & 10 & 11 \\
        \midrule
        1 & False & 0 & 0 & 0 & 0 & 0 \\
        2 & $A \land B$ & $A \cdot B$ & 0 & 0 & 0 & 1 \\
        3 & $\neg(A \Rightarrow B)$ & $A - AB$ & 0 & 0 & 1 & 0 \\
        \vdots & \vdots & \vdots & \vdots & \vdots & \vdots & \vdots\\
        14 & $A \Rightarrow B$ & $1 - A + AB$ & 1 & 1 & 0 & 1 \\
        15 & $\neg(A \land B)$ & $1 - AB$ & 1 & 1 & 1 & 0 \\
        16 & True & 1 & 1 & 1 & 1 & 1 \\
        \bottomrule
    \end{tabular}
\end{table}

\section{Formal Definition of Thermometer Encoding}
\label{apx:thermometer}
For a real-valued input $q \in \mathbb{R}$, let $z$ be the number of bits in the encoding, and $\vec{t} \in \mathbb{R}^z$ be the ordered Thermometer thresholds ($t_{i+1} > t_{i}$). The Thermometer Encoding $T \in \{0, 1\}^z$ of $q$ is given by:
\[ T(q) = (1_{q > t_1}, 1_{q > t_2}, \ldots, 1_{q > t_{z}}) \]

Here, \( 1_{q > t_i} \) is an indicator function, returning 1 if \( q > t_i \) and 0 otherwise.

\section{DiffLogicNet's Binary Logic Node}
\label{apx:difflogicnet}
Formally, for two input bit probabilities $A, B \in [0, 1]$, the output of each binary logic node during training is computed as:
\vspace{-1mm}
\[\sum_{i=1}^{16} \frac{e^{w_i}}{\sum_{j} e^{w_j}}  \cdot f_i(A, B)\]
\vspace{-4mm}

Here, $f_i$ denotes the $i$-th real-valued binary operator among the set of 16 distinct operators (a comprehensive list of which can be found in Appendix A, Table \ref{tab:difflogic}).
During inference, the network is discretized, with each binary logic node adopting the binary operator that has the highest associated weight.

\section{Out-of-Context Results}
\label{apx:ooc_dwn_fpga}

In our primary Table \ref{tab:fpga_results}, we reported FPGA results where input buffers and decompression were implemented, assuming data is streamed to the board using I/O pins (112 bits per cycle on the Zynq Z-7045 FPGA) and limited the clock speed to 200 MHz. These design choices were made to enable a fair comparison to prior work, particularly FINN, which adheres to similar constraints.

However, we noticed that other papers report results assuming data is already available on the board without accounting for streaming or decompression overheads (i.e., operating in out-of-context mode). To provide a comprehensive comparison, we present these out-of-context results in the Table \ref{tab:ooc_fpga_results_dwn}, following the methodology of other papers by setting the clock speed to the board's maximum of 250 MHz.

In this configuration, our model achieves an effective throughput of 250 million images per second. This is because our design is fully pipelined, taking only 4 ns per image during inference as the model is compact enough to process one image per clock cycle.

\begin{table*}[ht]
\vspace{-2mm}
\centering
\setlength{\tabcolsep}{4pt}
\caption{Implementation results for DWNs on a Z-7045 FPGA compiled in out-of-context mode with the clock set to an optimal 250 MHz (4 ns per cycle), the maximum frequency of the board.}
\label{tab:ooc_fpga_results_dwn}
\begin{tabular}{@{}llr@{\hspace{12pt}}r@{\hspace{12pt}}c@{\hspace{12pt}}r@{\hspace{12pt}}r@{\hspace{4pt}}}
\toprule
\multicolumn{1}{c}{\multirow{2}{*}{\textbf{Dataset}}} & 
  \multicolumn{1}{c}{\multirow{2}{*}{\textbf{Model}}} &
  \multirow{2}{*}{\shortstack[c]{\textbf{Test}\\\textbf{Accuracy\%}}}\hspace{-5pt} &
  \multirow{2}{*}{\shortstack[c]{\textbf{Parameter}\\\textbf{Size (KiB)}}}\hspace{-5pt} &
  \multirow{2}{*}{\shortstack[c]{\textbf{Time/Sample}}}\hspace{-5pt} &
  \multirow{2}{*}{\shortstack[c]{\textbf{Throughput}\\\textbf{(Samples/s)}}}\hspace{-5pt} &
  \multirow{2}{*}{\shortstack[c]{\textbf{LUTs}\\\textbf{(1000s)}}}\hspace{-5pt} \\
& & & & & & \\ \midrule
\multirow{4}{*}{MNIST}        
    & DWN \textit{($n$=2; lg)}     & 98.27 & \textbf{5.9} & 4ns & 250 M & 6.5 \\
    & DWN \textit{($n$=6; sm)}     & 97.80 & 11.7 & 4ns & 250 M & \textbf{2.1} \\
    & DWN \textit{($n$=6; lg)}     & 98.31 & 23.4  & 4ns & 250 M & 4.1 \\
    & DWN \textit{($n$=6; lg; +aug)} & \textbf{98.77} & 23.4  & 4ns & 250 M & 4.1 \\
\midrule
\multirow{2}{*}{FashionMNIST}
    & DWN \textit{($n$=2)} & \textbf{89.12} & \textbf{7.8} & 4ns & 250 M & 8.1 \\
    & DWN \textit{($n$=6)} & 89.01 & 31.3  & 4ns & 250 M & \textbf{6.2} \\
\midrule
\multirow{2}{*}{KWS} 
    & DWN \textit{($n$=2)} & 70.92 & \textbf{2.9} & 4ns & 250 M & 3.3 \\
    & DWN \textit{($n$=6)} & \textbf{71.52} & 12.5 & 4ns & 250 M & \textbf{3.3} \\
\midrule
\multirow{3}{*}{ToyADMOS/car} 
    & DWN \textit{($n$=2; sm)} & 86.68 & \textbf{0.9} & 4ns & 250 M & 1.0 \\
    & DWN \textit{($n$=6; sm)} & 86.93 & 3.1 & 4ns & 250 M & \textbf{0.8} \\
    & DWN \textit{($n$=6; lg)} & \textbf{89.03} & 28.1 & 4ns & 250 M & 5.5 \\
\midrule
\multirow{2}{*}{CIFAR-10} 
    & DWN \textit{($n$=2)} & 57.51 & \textbf{23.4} & 4ns & 250 M & 45.7 \\
    & DWN \textit{($n$=6)} & 57.42 & 62.5 & 4ns & 250 M & \textbf{16.7} \\
\bottomrule
\end{tabular}
\vspace{-3mm}
\end{table*}

\section{Bit-Packed Microcontroller Implementation Details}
\label{apx:microcontroller}

\begin{figure}[t]
\centerline{\includegraphics[width=1.0\columnwidth]{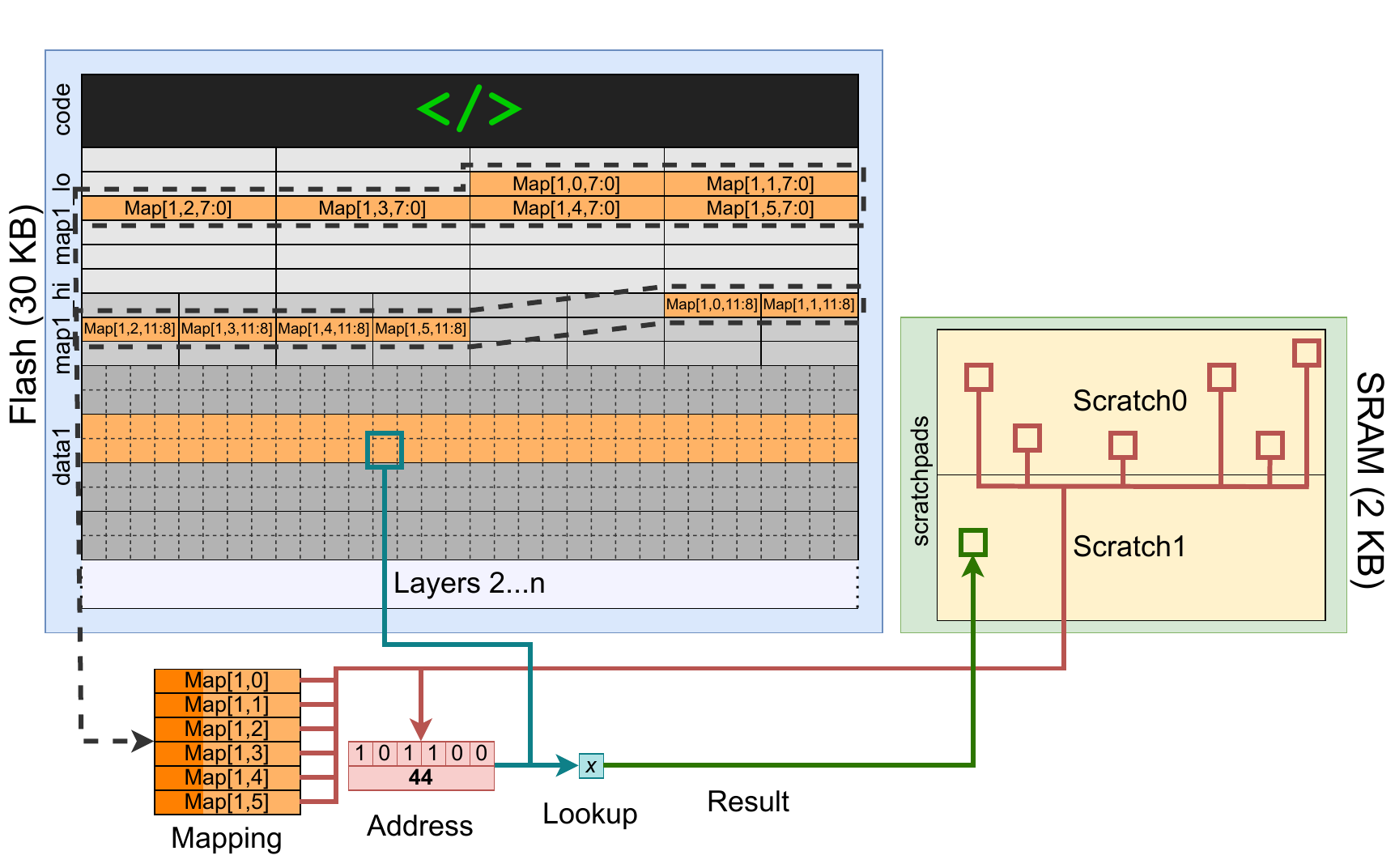}}
\caption{An overview of the data layout of a DWN model implemented on the Elegoo Nano. This microcontroller has very limited resources, which necessitates careful memory management.}
\label{fig:arduino_layout}
\end{figure}

Figure \ref{fig:arduino_layout} shows the memory layout for our bit-packed DWN inference model on the Elegoo Nano microcontroller. 
Flash memory is used to store program code, RAM node mapping information, and the contents of the RAM nodes themselves. Mapping arrays specify the indices of each input to each RAM node in a layer. These indices can be 8, 10, 12, or 16 bits, depending on the number of unique inputs to the layer.
In the case of an irregular bit width (10 or 12 bits), the map is further subdivided into \texttt{map\_lo}, which stores the low 8 bits, and \texttt{map\_hi}, which stores the remaining high bits. RAM node entries are packed, with 8 entries per byte.
For simplicity, this figure only shows mapping and data arrays for a single layer.

SRAM is divided into two large scratchpads, which layers alternate between. Layer $i$ of a model reads its input activations from scratchpad $(i-1) \bmod 2$ and writes its outputs to scratchpad $i \bmod 2$.

Bit-packing incurs a heavy runtime overhead. For instance, a single 6-input RAM node on a layer with 12-bit input mappings must (1) read the low byte of an input index from \texttt{map\_lo}, (2) read the high byte from \texttt{map\_hi}, (3) shift and mask the data read from \texttt{map\_hi} to isolate the relevant 4 bits, (4) shift and OR these bits with the byte read from \texttt{map\_lo}, (5) isolate and read the indicated bit from the input scratchpad, (6) repeat steps 1-5 five additional times, (7) construct an address from the six bits thus retrieved, (8) isolate and read the addressed bit from the data array, and lastly (9) set the result bit in the output scratchpad.

Bit-packing reduces the data footprint of models in flash memory by $\sim$4.5$\times$ and in SRAM by 8$\times$, which significantly increases the complexity of the models we can implement. Our unpacked microcontroller inference model is 9.5$\times$ faster than this packed implementation on average, but also 4.2\% less accurate.

All models on the Nano (for both DWNs and XGBoost) communicate with a host PC over a 1 Mbps serial connection, which they use to receive input samples and send back predictions.
This was chosen to minimize data movement overhead while maintaining reliability; while a 2 Mbps connection is theoretically possible, we found that it was unstable in practice.


\section{Estimating NAND2 Equivalents for Popcount Circuits}
\label{apx:popcount}

\begin{figure}[htbp]
\centerline{\includegraphics[width=0.8\columnwidth]{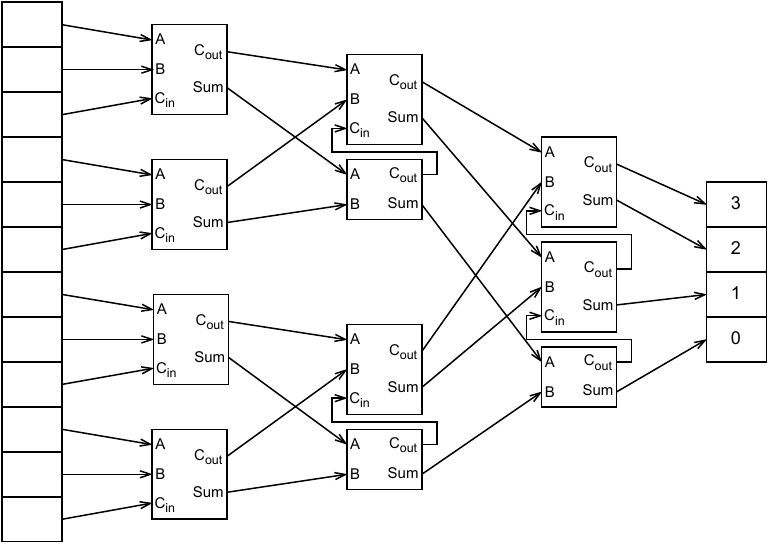}}
\caption{A 12:4 popcount tree composed of 8 full adders and 3 half adders.}
\label{fig:adder_tree}
\end{figure}

To obtain NAND2 equivalent areas of DiffLogicNet models, we need a way to estimate the area of an $N$-input popcount.
Let's first consider the number of half adders and full adders that are required. As shown in Figure~\ref{fig:adder_tree}, we can construct a popcount circuit by first passing trios of 1-bit inputs into full adders, which results in $\lceil \frac{N}{3} \rceil$ partial sums which are each 2 bits wide. We then pass these signals into a reduction tree of adders. The $i$th layer of this tree requires approximately $\frac{n}{3} 2^{-i}$ multi-bit adders which are each $i$+1 bits wide.
These multi-bit adders are in turn composed of one half adder and $i$ full adders.
Therefore, including $\frac{N}{3}$ full adders from the first layer, this popcount circuit requires approximately
\[
    \frac{N}{3} + \sum_{i=1}^{\log_{2}(N/3)}i\left(\frac{N}{3}2^{-i}\right) = N - \log_{2}\left(\frac{N}{3}\right) - 2
\]
full adders, and
\[
    \sum_{i=1}^{\log_{2}(N/3)}\left(\frac{N}{3}2^{-i}\right) = \frac{N}{3} - 1
\]
half adders.

A naive implementation of this circuit requires 5 NAND2 gates per half adder and 9 gates per full adder.
However, we can obtain a lower and more accurate estimate by looking at the transistor-level optimizations performed in standard cell libraries. For instance, the open-source SkyWater SKY130 PDK~\cite{skywater} implements the smallest NAND2 gate using 4 transistors, a half-adder using 14 transistors, and a full-adder using 28 transistors.
Therefore, we approximate a half adder as 3.5 NAND2 gates and a full adder as 7 NAND2 gates.

Our final NAND2 equivalent estimate for an $N$-input popcount is:
\[
    7\left(N - \log_{2}\left(\frac{N}{3}\right) - 2\right) + 3.5\left(\frac{N}{3} - 1\right) \approx 8.167N - 7\log_{2}(N) - 6.405
\]

\section{Expanded Details on Tabular Data Comparisons for Tiny Models}

Table~\ref{tab:tiny_tabular_full} provides additional details for our comparison between DWNs and tiny tabular models.
For each DiffLogicNet model, we separately break out the NAND2 equivalent areas for the logic and the popcount circuit.
For the Tiny Classifier models, we include additional, smaller configurations with 50, 100, or 200 NAND gates from their paper.

\begin{table*}[htbp]
\centering
\caption{Expanded Table \ref{tab:tiny_tabular} detailing DiffLogicNets' Logic NAND count and Total NAND count, alongside additional model sizes for Tiny Classifiers.}
\label{tab:tiny_tabular_full}
\begin{tabular}{@{}lcrrcrrccccc@{}}
\toprule
\multirow{3}{*}{\textbf{Dataset}} &
  \multicolumn{2}{c}{\textbf{DWN}} &
  \textbf{} &
  \multicolumn{3}{c}{\textbf{DiffLogicNet}} &
  \multicolumn{1}{l}{} &
  \multicolumn{4}{c}{\textbf{Tiny Classifier \cite{tinyclassifier}}} \\ \cmidrule(lr){2-3} \cmidrule(lr){5-7} \cmidrule(l){9-12} 
 &
  \multirow{2}{*}{\textbf{Acc.}} &
  \multirow{2}{*}{\textbf{NAND}} &
  \textbf{} &
  \multirow{2}{*}{\textbf{Acc.}} &
  \textbf{Logic} &
  \textbf{Total} &
  \multicolumn{1}{l}{} &
  \textbf{50 NAND} &
  \textbf{100 NAND} &
  \textbf{200 NAND} &
  \textbf{300 NAND} \\
 &
   &
   &
  \textbf{} &
   &
  \textbf{NAND} &
  \textbf{NAND} &
  \multicolumn{1}{l}{} &
  \textbf{Acc.} &
  \textbf{Acc.} &
  \textbf{Acc.} &
  \textbf{Acc.} \\ \midrule
phoneme    & 0.857 & 168 &  & 0.832 & 394  & 836  &                      & 0.700 & 0.700 & 0.780 & 0.790 \\
skin-seg   & 0.989 & 88  &  & 0.982 & 327  & 610   &                      & 0.900 & 0.890 & 0.950 & 0.930 \\
higgs      & 0.678 & 94  &  & 0.675 & 232  & 658   &                      & 0.600 & 0.520 & 0.620 & 0.620 \\
australian & 0.885 & 7   &  & 0.877 & 175  & 379   &                      & 0.800 & 0.830 & 0.850 & 0.850 \\
nomao      & 0.935 & 87  &  & 0.935 & 2612 & 4955  &                      & 0.800 & 0.750 & 0.780 & 0.800 \\
segment    & 0.994 & 71  &  & 0.994 & 248  & 610   &                      & 0.500 & 0.790 & 0.790 & 0.950 \\
miniboone  & 0.901 & 60  &  & 0.901 & 249  & 619  &                      & 0.730 & 0.750 & 0.780 & 0.820 \\
christine  & 0.706 & 53  &  & 0.683 & 3147 & 16432 & \multicolumn{1}{l}{} & 0.550 & 0.540 & 0.580 & 0.590 \\
jasmine    & 0.806 & 51  &  & 0.767 & 689  & 1816  &                      & 0.500 & 0.710 & 0.710 & 0.720 \\
sylvine    & 0.922 & 44  &  & 0.857 & 202  & 501   &                      & 0.890 & 0.890 & 0.890 & 0.890 \\
blood      & 0.784 & 49  &  & 0.784 & 82   & 365   &                      & 0.600 & 0.600 & 0.620 & 0.630 \\
\bottomrule 
\end{tabular}
\end{table*}

\section{List of Datasets Used}
\label{apx:datasets}

See Table~\ref{tab:datasets}.

\begin{table*}[htbp]
\centering
\caption{List of all datasets used in this paper.}
\label{tab:datasets}
\begin{tabular}{@{}lll@{}}
\toprule
\textbf{Dataset Name} & \textbf{Source} & \textbf{Notes} \\
\midrule
Iris & \cite{misc_iris_53} & Extremely simple dataset; used in Figure 1 for illustrative purposes. \\
MNIST & \cite{mnist} & \\
FashionMNIST & \cite{xiao2017/online} & Designed to be the same size as MNIST but more difficult. \\
KWS & \cite{speech_commands} & Subset of dataset; part of MLPerf Tiny~\cite{mlperf}. \\
ToyADMOS/car & \cite{toyadmos} & Subset of dataset; part of MLPerf Tiny~\cite{mlperf}. \\
CIFAR10 & \cite{cifar10} & Part of MLPerf Tiny~\cite{mlperf}. \\
JSC & \cite{duarte2018fast} & \\
phoneme & \cite{phoneme} & \\
skin-seg & \cite{misc_skin_segmentation_229} & \\
higgs & \cite{misc_higgs_280} & \\
australian & \cite{misc_statlog_143} & \\
nomao & \cite{misc_nomao_227} & \\
segment & \cite{misc_image_segmentation_50} & Binarized version of original dataset \\
miniboone & \cite{misc_miniboone_particle_identification_199} & \\
christine & \cite{chalearn_competition} & Synthetic dataset, created for competition. \\
jasmine & \cite{chalearn_competition} & Synthetic dataset, created for competition. \\
sylvine & \cite{chalearn_competition} & Synthetic dataset, created for competition. \\
blood & \cite{misc_blood_transfusion_service_center_176} & \\
\bottomrule
\end{tabular}
\end{table*}

\section{Ablation Studies}

See Tables \ref{tab:ablation_efd_lm}, \ref{tab:ablation_efd_lm_change}, \ref{tab:ablation_spectral} and \ref{tab:ablation_reduction}.

\begin{table}[]
\centering
\caption{Ablation Studies comparing Finite Difference (FD) (a minimal DWN) to Extended Finite Difference (EFD) and Learnable Mapping (LM). $^*$These results are from models with only one layer, so input derivatives are not needed if LM is not employed. Therefore, FD and EFD behave the same when LM is not used.}
\label{tab:ablation_efd_lm}
\begin{tabular}{@{}lcccc@{}}
\toprule
\textbf{Dataset}      & \textbf{FD (Minimal DWN)} & \textbf{+ EFD}   & \textbf{+ LM}    & \textbf{+ EFD + LM} \\ \midrule
MNIST        & 96.15\%          & 96.59\% & 98.30\% & 98.31\%               \\
FashionMNIST & 85.74\%          & 86.88\% & 87.94\% & 89.01\%               \\
KWS          & 52.33\%*          & 52.33\%* & 70.24\% & 71.52\%               \\
ToyADMOS/car & 87.73\%          & 88.02\% & 88.52\% & 89.03\%               \\
CIFAR-10     & 48.37\%*          & 48.37\%* & 55.36\% & 57.42\%               \\ \bottomrule
\end{tabular}
\end{table}

\begin{table}[]
\centering
\caption{Gains in accuracy observed by each method in Table \ref{tab:ablation_efd_lm}. $^*$These results are from models with only one layer, so input derivatives are not needed if LM is not employed. Therefore, FD and EFD behave the same when LM is not used.}
\label{tab:ablation_efd_lm_change}
\begin{tabular}{@{}lrrr@{}}
\toprule
\textbf{Dataset}      & \textbf{+ EFD}  & \textbf{+ LM}    & \textbf{+ EFD + LM} \\ \midrule
MNIST        & 0.44\% & 2.15\%  & 2.16\%     \\
FashionMNIST & 1.14\% & 2.20\%  & 3.27\%     \\
KWS          & 0.00\%* & 17.91\% & 19.19\%    \\
ToyADMOS/car & 0.29\% & 0.79\%  & 1.30\%     \\
CIFAR-10     & 0.00\%* & 6.99\%  & 9.05\%     \\ \bottomrule
\end{tabular}
\end{table}

\begin{table}[]
\centering
\caption{Ablation Studies comparing a DWN with and without Spectral Normalization (SN)}
\label{tab:ablation_spectral}
\begin{tabular}{@{}lrrr@{}}
\toprule
\textbf{Dataset}      & \multicolumn{1}{c}{\textbf{DWN}} & \multicolumn{1}{c}{\textbf{+ SN}} & \multicolumn{1}{c}{\textbf{Change}} \\ \midrule
MNIST        & 97.82\%                 & 97.88\%                  & +0.06\%                     \\
FashionMNIST & 87.10\%                 & 88.16\%                  & +1.06\%                     \\
KWS          & 67.17\%                 & 69.60\%                  & +2.43\%                     \\
ToyADMOS/car & 88.04\%                 & 88.68\%                  & +0.64\%                     \\
phoneme      & 89.55\%                 & 89.50\%                  & -0.05\%                    \\
skin-seg     & 99.65\%                 & 99.83\%                  & +0.18\%                     \\
higgs        & 68.51\%                 & 72.42\%                  & +3.91\%                     \\ \bottomrule
\end{tabular}
\end{table}

\begin{table}[]
\centering
\caption{Ablation Studies comparing the NAND2 gate count of DWN for Ultra-Low-Cost Chips with and without Learnable Reduction (LR). The results for DWN without LR are divided into three columns: the NAND count used by the LUTs (Logic), the NAND count used by the Popcount, and the total (Logic + Popcount). The DWN + LR column shows directly the total model size as no Popcount is needed when employing LR.}
\label{tab:ablation_reduction}
\begin{tabular}{@{}lrrrrc@{}}
\toprule
\multirow{2}{*}{\textbf{Dataset}} & \multicolumn{3}{c}{\textbf{DWN}}                                                              & \multicolumn{1}{c}{\multirow{2}{*}{\textbf{DWN + LR}}} & \multicolumn{1}{c}{\textbf{Circuit}}   \\ 
                         & \multicolumn{1}{c}{\textbf{Logic}} & \multicolumn{1}{c}{\textbf{Popcount}} & \multicolumn{1}{c}{\textbf{Total}} & \multicolumn{1}{c}{}                          & \multicolumn{1}{c}{\textbf{Reduction}} \\ \midrule
phoneme                  & 146                       & 111                          & 257                       & 168                                           & 1.53x                          \\
skin-seg                 & 189                       & 157                          & 346                       & 88                                            & 3.93x                          \\
higgs                    & 90                        & 141                          & 231                       & 94                                            & 2.46x                          \\
australian               & 12                        & 24                           & 36                        & 7                                             & 5.14x                          \\
nomao                    & 84                        & 126                          & 210                       & 87                                            & 2.41x                          \\
segment                  & 49                        & 111                          & 160                       & 71                                            & 2.25x                          \\
miniboone                & 55                        & 66                           & 121                       & 60                                            & 2.02x                          \\
christine                & 45                        & 52                           & 97                        & 53                                            & 1.83x                          \\
jasmine                  & 48                        & 66                           & 114                       & 51                                            & 2.24x                          \\
sylvine                  & 63                        & 96                           & 159                       & 44                                            & 3.61x                          \\
blood                    & 60                        & 96                           & 156                       & 49                                            & 3.18x                          \\ \bottomrule
\end{tabular}
\end{table}

\section{Model Configurations}
\label{apx:modelconfig}

See Tables \ref{tab:fpgaconfig}, \ref{tab:arduinoconfig}, \ref{tab:tinyconfig} and \ref{tab:tabularconfig}. 

\textbf{z} indicates the number of thermometer bits per feature utilized in the binary encoding, \textbf{tau} the softmax temperature utilized after the popcount during training, and \textbf{BS} the batch size.

\begin{table}[]
\centering
\caption{DWN model configurations for Table \ref{tab:fpga_results} and Table \ref{tab:lutmodels}. All models were trained for a total of 100 Epochs.}
\label{tab:fpgaconfig}
\begin{tabular}{@{}llccccc@{}}
\toprule
\textbf{Dataset}                       & \multicolumn{1}{c}{\textbf{Model}} & \textbf{z} & \textbf{Layers} & \textbf{tau} & \textbf{BS} & \textbf{Learning Rate}                          \\ \midrule
\multirow{4}{*}{MNIST}        & DWN (n=2; lg)             & 3                  & 2x 6000    & 1/0.071      & 128        & 1e-2(30), 1e-3(30), 1e-4(30), 1e-5(10)    \\
                              & DWN (n=6; sm)             & 1                  & 1000, 500  & 1/0.245      & 128        & 1e-2(30), 1e-3(30), 1e-4(30), 1e-5(10)    \\
                              & DWN (n=6; lg)             & 3                  & 2000, 1000 & 1/0.173      & 128        & 1e-2(30), 1e-3(30), 1e-4(30), 1e-5(10)    \\
                              & DWN (n=6; lg + aug)       & 3                  & 2000, 1000 & 1/0.173      & 128        & 1e-2(30), 1e-3(30), 1e-4(30), 1e-5(10)    \\ \midrule
\multirow{2}{*}{FashionMNIST} & DWN (n=2)                 & 7                  & 2x 8000    & 1/0.061      & 128        & 1e-2(30), 1e-3(30), 1e-4(30), 1e-5(10)    \\
                              & DWN (n=6)                 & 7                  & 2x 2000    & 1/0.122      & 128        & 1e-2(30), 1e-3(30), 1e-4(30), 1e-5(10)    \\ \midrule
\multirow{2}{*}{KWS}          & DWN (n=2)                 & 8                  & 1x 3000     & 1/0.1        & 100        & 1e-2(30), 1e-3(30), 1e-4(30), 1e-5(10)    \\
                              & DWN (n=6)                 & 8                  & 1x 1600    & 1/0.1        & 100        & 1e-2(30), 1e-3(30), 1e-4(30), 1e-5(10)    \\ \midrule
\multirow{3}{*}{ToyADMOS/car} & DWN (n=2; sm)             & 2                  & 2x 900     & 1/0.1        & 100        & 1e-2(30), 1e-3(30), 1e-4(30), 1e-5(10)    \\
                              & DWN (n=6; sm)             & 2                  & 1x 400     & 1/0.1        & 100        & 1e-2(30), 1e-3(30), 1e-4(30), 1e-5(10)    \\
                              & DWN (n=6; lg)             & 3                  & 2x 1800    & 1/0.058      & 100        & 1e-2(30), 1e-3(30), 1e-4(30), 1e-5(10)    \\ \midrule
\multirow{2}{*}{CIFAR-10}     & DWN (n=2)                 & 10                 & 2x 24000   & 1/0.03       & 100        & 1e-2(30), 1e-3(30), 1e-4(30), 1e-5(10)    \\
                              & DWN (n=6)                 & 10                 & 8000       & 1/0.03       & 100        & 1e-2(30), 1e-3(30), 1e-4(30), 1e-5(10)    \\ \midrule
\multirow{4}{*}{JSC}          & DWN (n-6; sm)             & 200                & 1x 10      & 1/0.7        & 100        & 1e-2(14), 1e-3(14), 1e-4(4)    \\
                              & DWN (n-6; sm)             & 200                & 1x 50      & 1/0.3        & 100        & 1e-2(14), 1e-3(14), 1e-4(4)    \\
                              & DWN (n-6; md)             & 200                & 1x 360    & 1/0.1       & 100        & 1e-2(14), 1e-3(14), 1e-4(4)    \\
                              & DWN (n-6; lg)             & 200                & 1x 2400    & 1/0.03       & 100        & 1e-2(14), 1e-3(14), 1e-4(4)    \\ \bottomrule
\end{tabular}
\end{table}

\begin{table*}[]
\centering
\caption{DWN model configurations for Table \ref{tab:arduino}.}
\label{tab:arduinoconfig}
\begin{tabular}{@{}llcccccc@{}}
\toprule
\textbf{Model} & \textbf{Dataset} & \textbf{z} & \textbf{Layers} & \textbf{tau} & \textbf{BS} & \textbf{Learning Rate}                 & \textbf{Epochs} \\ \midrule
\multirowcell{7}[0pt][l]{Bit-Packed} & MNIST            & 3                           & 1000, 500       & 1/0.077               & 128                 & 1e-2(30), 1e-3(30), 1e-4(30), 1e-5(10) & 100             \\
& FashionMNIST     & 3                           & 1000, 500       & 1/0.077               & 128                 & 1e-2(30), 1e-3(30), 1e-4(30), 1e-5(10) & 100             \\
& KWS              & 3                           & 1000, 500       & 1/0.077               & 128                 & 1e-2(30), 1e-3(30), 1e-4(30), 1e-5(10) & 100             \\
& ToyADMOS/car     & 3                           & 1000, 500       & 1/0.077               & 128                 & 1e-2(30), 1e-3(30), 1e-4(30), 1e-5(10) & 100             \\
& phoneme          & 128                         & 1000, 500       & 1/0.077               & 256                 & 1e-2(30), 1e-3(30), 1e-4(30), 1e-5(10) & 100             \\
& skin-seg         & 128                         & 1000, 500       & 1/0.077               & 256                 & 1e-2(30), 1e-3(30), 1e-4(30), 1e-5(10) & 100             \\
& higgs            & 128                         & 1000, 500       & 1/0.077               & 128                 & 1e-2(30), 1e-3(30), 1e-4(30), 1e-5(10) & 100             \\ \midrule
\multirowcell{7}[0pt][l]{Unpacked} & MNIST & 32 & 220, 110 & 1/0.165 & 128 & 1e-2(30), 1e-3(30), 1e-4(30), 1e-5(10) & 100 \\
& FashionMNIST & 32 & 220, 110 & 1/0.165 & 128 & 1e-2(30), 1e-3(30), 1e-4(30), 1e-5(10) & 100 \\
& KWS &  32 & 220, 110 & 1/0.165 & 128 & 1e-2(30), 1e-3(30), 1e-4(30), 1e-5(10) & 100 \\
& ToyADMOS & 32 & 220, 110 & 1/0.165 & 128 & 1e-2(30), 1e-3(30), 1e-4(30), 1e-5(10) & 100 \\
& phoneme & 255 & 80, 40 & 1/0.274 & 256 & 1e-2(30), 1e-3(30), 1e-4(30), 1e-5(10) & 100 \\
& skin-seg & 255 & 80, 40 & 1/0.274 & 256 & 1e-2(30), 1e-3(30), 1e-4(30), 1e-5(10) & 100 \\
& higgs & 255 & 90, 90 & 1/0.183 & 128 & 1e-2(30), 1e-3(30), 1e-4(30), 1e-5(10) & 100 \\
\bottomrule
\end{tabular}
\end{table*}

\begin{table}[]
\centering
\caption{DWN model configurations for Table \ref{tab:tiny_tabular}.}
\label{tab:tinyconfig}
\begin{tabular}{@{}lcccccc@{}}
\toprule
\textbf{Dataset} & \textbf{z} & \textbf{Layers}        & \textbf{tau} & \multicolumn{1}{c}{\textbf{BS}} & \textbf{Learning Rate}          & \textbf{Epochs} \\ \midrule
phoneme          & 200                         & 64, 32, 16, 8, 4, 2, 1 & 1/0.03                & 32                                      & 1e-2 (80), 1e-3 (80), 1e-4 (40) & 200             \\
skin-seg         & 200                         & 64, 32, 16, 8, 4, 2, 1 & 1/0.001               & 32                                      & 1e-2 (80), 1e-3 (80), 1e-4 (40) & 200             \\
higgs            & 200                         & 64, 32, 16, 8, 4, 2, 1 & 1/0.001               & 32                                      & 1e-2 (80), 1e-3 (80), 1e-4 (40) & 200             \\
australian       & 200                         & 4, 2, 1                & 1/0.03                & 32                                      & 1e-2 (80), 1e-3 (80), 1e-4 (40) & 200             \\
nomao            & 200                         & 64, 32, 16, 8, 4, 2, 1 & 1/0.03                & 32                                      & 1e-2 (80), 1e-3 (80), 1e-4 (40) & 200             \\
segment          & 200                         & 64, 32, 16, 8, 4, 2, 1 & 1/0.03                & 32                                      & 1e-2 (80), 1e-3 (80), 1e-4 (40) & 200             \\
miniboone        & 200                         & 64, 32, 16, 8, 4, 2, 1 & 1/0.001               & 32                                      & 1e-2 (80), 1e-3 (80), 1e-4 (40) & 200             \\
christine        & 200                         & 64, 32, 16, 8, 4, 2, 1 & 1/0.03                & 32                                      & 1e-2 (80), 1e-3 (80), 1e-4 (40) & 200             \\
jasmine          & 200                         & 64, 32, 16, 8, 4, 2, 1 & 1/0.03                & 32                                      & 1e-2 (80), 1e-3 (80), 1e-4 (40) & 200             \\
sylvine          & 200                         & 64, 32, 16, 8, 4, 2, 1 & 1/0.03                & 32                                      & 1e-2 (80), 1e-3 (80), 1e-4 (40) & 200             \\
blood            & 200                         & 8, 4, 2, 1             & 1/0.03                & 32                                      & 1e-2 (80), 1e-3 (80), 1e-4 (40) & 200             \\ \bottomrule
\end{tabular}
\end{table}

\begin{table}[]
\centering
\caption{DWN model configurations for Table \ref{tab:big_tabular}.}
\label{tab:tabularconfig}
\begin{tabular}{@{}lcccrcc@{}}
\toprule
\textbf{Dataset} & \textbf{z} & \textbf{Layers} & \textbf{tau} & \multicolumn{1}{c}{\textbf{BS}} & \textbf{Learning Rate}          & \textbf{Epochs} \\ \midrule
phoneme          & 200                         & 3x 21000            & 1/0.03                & 32                                      & 1e-2 (80), 1e-3 (80), 1e-4 (40) & 200             \\
skin-seg         & 200                         & 5x 10000            & 1/0.001               & 32                                      & 1e-2 (80), 1e-3 (80), 1e-4 (40) & 200             \\
higgs            & 200                         & 5x 22000            & 1/0.001               & 32                                      & 1e-2 (80), 1e-3 (80), 1e-4 (40) & 200             \\
australian       & 200                         & 3x 12000            & 1/0.03                & 32                                      & 1e-2 (80), 1e-3 (80), 1e-4 (40) & 200             \\
nomao            & 200                         & 3x 24000            & 1/0.03                & 32                                      & 1e-2 (80), 1e-3 (80), 1e-4 (40) & 200             \\
segment          & 200                         & 3x 7500             & 1/0.03                & 32                                      & 1e-2 (80), 1e-3 (80), 1e-4 (40) & 200             \\
miniboone        & 200                         & 5x 22000            & 1/0.001               & 32                                      & 1e-2 (80), 1e-3 (80), 1e-4 (40) & 200             \\
christine        & 20                          & 3x 26000            & 1/0.03                & 32                                      & 1e-2 (80), 1e-3 (80), 1e-4 (40) & 200             \\
jasmine          & 200                         & 3x 20000            & 1/0.03                & 32                                      & 1e-2 (80), 1e-3 (80), 1e-4 (40) & 200             \\
sylvine          & 200                         & 3x 17000            & 1/0.03                & 32                                      & 1e-2 (80), 1e-3 (80), 1e-4 (40) & 200             \\
blood            & 200                         & 3x 13000            & 1/0.03                & 32                                      & 1e-2 (80), 1e-3 (80), 1e-4 (40) & 200             \\ \bottomrule
\end{tabular}
\end{table}